\pdfoutput=1
\documentclass[11pt,a4paper]{article}

\usepackage[margin=1in]{geometry}
\usepackage{amsmath,amsfonts,amssymb}
\usepackage{graphicx}
\usepackage[authoryear]{natbib}
\usepackage{booktabs}
\usepackage{multirow}
\usepackage{array}
\usepackage{xcolor}
\usepackage{tabularx}
\usepackage{algorithm}
\usepackage{algpseudocode}
\usepackage{float}
\usepackage{authblk}
\usepackage[colorlinks=true,citecolor=blue,linkcolor=blue,urlcolor=blue]{hyperref}

\title{Unifying Active Learning and Semi-Supervised Learning for Medical Image Segmentation}

\author[1]{Bahram Jafrasteh\thanks{Corresponding author. Email: \texttt{baj4003@med.cornell.edu}}}
\author[1,2]{Cheng Wan}
\author[1]{Heejong Kim}
\author[1]{Johannes C. Paetzold}
\author[1]{Qingyu Zhao\thanks{Email: \texttt{qiz4006@med.cornell.edu}}}
\affil[1]{Department of Radiology, Weill Cornell Medicine, New York, NY, USA}
\affil[2]{Department of Electrical and Computer Engineering, Cornell University, Ithaca, NY, USA}
\date{}

\begin{document}
\maketitle

\begin{abstract}
In practical settings, medical image segmentation models are often developed with limited annotated data rather than fully labeled datasets. Training frequently begins in ultra-low labeled regimes where only a small number of volumes are annotated. In such scenarios, practitioners must simultaneously decide which cases to annotate and how to best use the remaining unlabeled data. Although active learning (AL) and semi-supervised learning (SSL) both target annotation scarcity, they are typically designed and optimized independently, resulting in objective mismatch and unstable training during early-stage “cold start” conditions.
We propose RegAL, a unified active semi-supervised framework governed by a shared topology-aware Pareto optimization that couples sample acquisition with unlabeled data utilization. RegAL evaluates images along three complementary axes, voxel-wise uncertainty, feature diversity, and a novel topological consistency metric, to select anatomically informative edge cases for annotation. On the other hand, the same criteria are used to identify geometrically stable atlas candidates for diffeomorphic registration-guided augmentation to train a self-supervised Mean Teacher segmentation network.
Across BraTS 2021, dHCP, and ProstateX, RegAL remains stable with few labeled volumes and consistently outperforms state-of-the-art AL, SSL, and active semi-supervised baselines across Dice and boundary-distance (ASD, HD95) metrics under extreme annotation scarcity.
\end{abstract}

\noindent\textbf{Keywords:} Active Learning, Semi-Supervised Learning, Medical Image Segmentation, Pareto Optimization, Image Registration

\section{Introduction}

Deep neural networks, spanning convolutional networks like 3D U-Net \citep{cciccek20163d} and transformer-based models \citep{hatamizadeh2022unetr}, have achieved remarkable performance in medical image segmentation. However, the success of these models is fundamentally based on large, densely annotated datasets, which are costly, time-consuming, and often infeasible to obtain in many clinical research settings.

In practice, medical imaging studies rarely begin with hundreds of labeled volumes. Instead, they typically start with a newly collected or institution-specific dataset containing no annotations, or at best a handful of manually labeled cases. Before large-scale annotation efforts are justified, researchers must demonstrate preliminary feasibility. This creates a pervasive yet under-addressed challenge: the \textit{ultra-low labeled regime}, where only a few annotated volumes are available. In this regime, standard supervised training is prone to overfitting and often fails catastrophically on small or ambiguous structures, a phenomenon referred to as the \textit{cold-start problem}. Despite its practical importance, relatively little work directly targets this cold-start scenario.

Two major research paradigms have emerged to address limited annotation settings. Active Learning (AL) aims to reduce annotation cost by selecting the most informative samples from an unlabeled pool for human annotation \citep{sener2018active, huang2010active}. Typical strategies rely on uncertainty estimation \citep{gal2016dropout}, diversity sampling, or core-set formulations to prioritize cases expected to maximally improve the model. Semi-Supervised Learning (SSL), in contrast, seeks to use large amounts of unlabeled data during training. Consistency regularization, Mean Teacher frameworks \citep{tarvainen2017mean}, and pseudo-labeling methods \citep{sohn2020fixmatch} attempt to extract supervisory signals from unlabeled volumes to improve representation learning.

While both AL and SSL target the same fundamental bottleneck, annotation scarcity, they are typically studied in isolation. While AL focuses on which samples to label under the assumption of supervised retraining, SSL focuses on how to use unlabeled data given a fixed labeled set. 
% AL focuses on which samples to label, assuming supervised training thereafter. SSL focuses on how to train with unlabeled data, assuming a pre-fixed labeled set. 
In real-world clinical pipelines, however, data acquisition and model training are intertwined:
% these two processes are inherently intertwined: 
practitioners iteratively annotate a few cases, retrain models, and attempt to exploit the remaining unlabeled data. Treating AL and SSL as independent modules overlooks this structural coupling.
% ignores this natural coupling.

\begin{figure}[t!]
    \centering
\includegraphics[width=\textwidth]{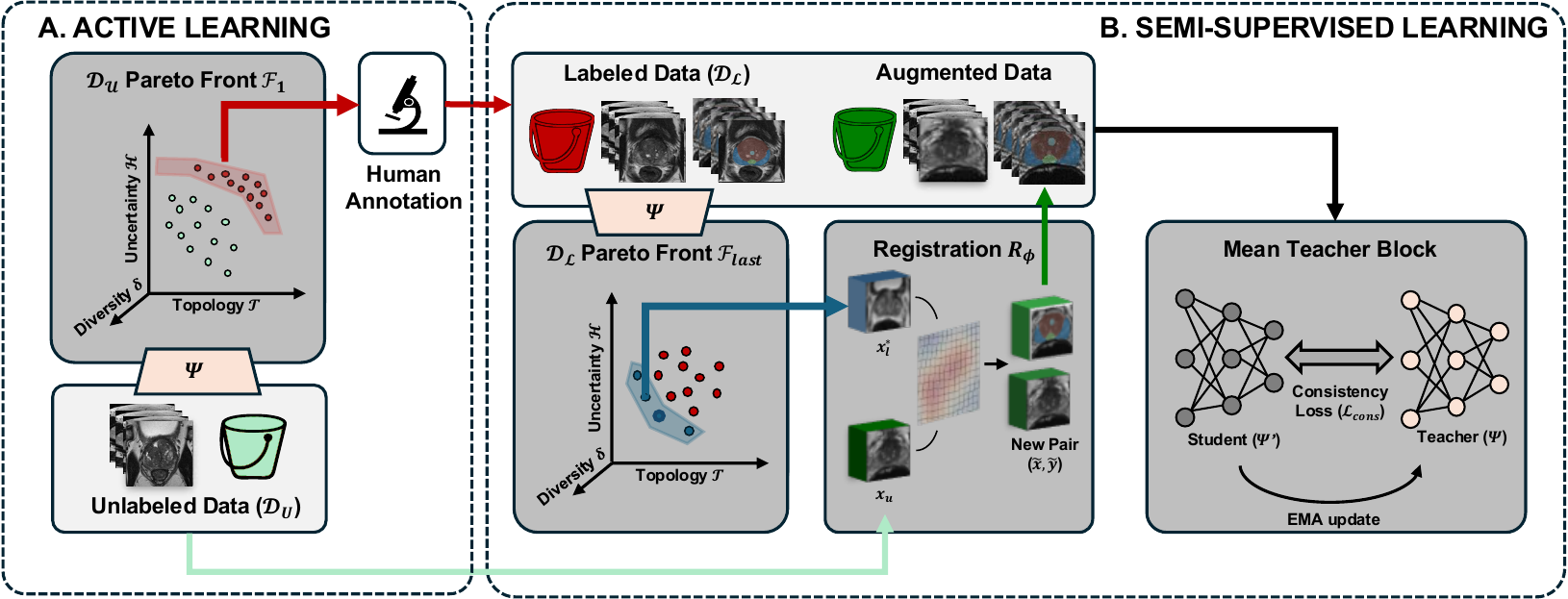}
\caption{The RegAL framework iterates between AL and SSL governed by Pareto optimization:
\textbf{A. AL:} At the start of a cycle, the unlabeled pool is evaluated with uncertainty, topology, and feature diversity, computed from latent features extracted by the Teacher network $\Psi$ from the previous cycle. A Pareto optimization ranks these samples, and challenging "edge cases" are selected for human annotation and added to the labeled set.
\textbf{B. SSL:} A Mean Teacher architecture is trained for multiple epochs. For each unlabeled $x_u$, the current labeled set is dynamically ranked using $\Psi$ from the previous epoch under the same Pareto criteria. An optimal atlas $x_l^*$ is diffeomorphically registered to $x_u$ to generate synthetic image–label pairs. The Student network $\Psi'$ is optimized with both real and registration-augmented data, together with a consistency loss aligning it to the Teacher $\Psi$, whose weights are updated via EMA.}
    \label{fig:fig1}
\end{figure}

Several recent works have explored Active Semi-Supervised Learning (ASSL) \citep{rangnekar2023semantic, wen2025active} to combine these paradigms. However, existing approaches typically stack AL and SSL sequentially using separate heuristics without a unifying principle. As a result, the two components operate under different objectives and assumptions, resulting in high vulnerability in the ultra-low labeled regime. 
We therefore argue that ASSL requires \textbf{a coherent, shared principle} governing both sample acquisition and unlabeled data utilization. Rather than treating AL and SSL as separate algorithmic blocks, they should be viewed as two directions of the same decision process: AL selects challenging edge cases to expand the knowledge boundary, while SSL identifies reliable structural cores to stabilize representation learning. Following this notion, we propose \textbf{RegAL} (Fig. \ref{fig:fig1}), a unified ASSL medical image segmentation framework based on \textit{topology-aware Pareto optimization}. RegAL has three novelties:

\begin{enumerate}
    \item \textbf{A unified Pareto-driven ASSL framework.} We introduce a bidirectional multi-objective Pareto optimization that governs both AL and SSL. The same principled selection mechanism is applied to acquire informative edge cases for AL and to identify stable structural cores for SSL, eliminating the objective mismatch present in prior ASSL pipelines.

    \item \textbf{Topology-aware AL.} In the AL phase, we jointly evaluate unlabeled samples using voxel-wise uncertainty ($\mathcal{H}$), feature diversity ($\delta$), and a novel topological consistency metric ($\mathcal{T}$).  Rather than selecting globally uncertain cases alone, RegAL identifies anatomically informative edge cases that are both difficult to segment and tend to result in structurally implausible predictions (e.g., disconnected components). Prioritizing these samples for annotation maximizes the value of each labeling effort and accelerates the learning of topologically accurate segmentation models. 
    %that explicitly detects structural fragmentation in predicted segmentations. Unlike global uncertainty measures ($\mathcal{H}$) that suffer from signal dilution in 3D volumes, our topology-aware objective captures anatomically implausible disconnected predictions, enabling the selection of structurally challenging samples in low data regimes.
    
    \item \textbf{Registration-guided SSL.} We propose a registration-guided augmentation strategy. For each unlabeled sample, a dynamically selected labeled atlas is warped via diffeomorphic transformations \citep{balakrishnan2019voxelmorph} to create topologically valid synthetic training data. This practice stabilizes training in extreme cold-start scenarios.
\end{enumerate}

Together, these contributions establish a coherent and topology-aware ASSL framework specifically designed for low labeled medical segmentation.

\section{Related Work}

Modern medical image segmentation is dominated by encoder-decoder convolutional networks such as the 3D U-Net \citep{cciccek20163d} and self-configuring pipelines like nnU-Net \citep{isensee2021nnunet}, as well as transformer-based architectures \citep{hatamizadeh2022unetr}. These models achieve strong accuracy when trained on large, densely annotated datasets, but their reliance on extensive voxel-level supervision motivates the annotation-efficient learning paradigms reviewed below.

\subsection{Active Learning in Medical Image Segmentation}
AL aims to minimize annotation cost by strategically selecting the most informative unlabeled samples for human review \citep{huang2010active, budd2021survey}. In medical imaging, AL strategies predominantly fall into two categories: uncertainty-based and diversity-based sampling. Uncertainty approaches, such as those using Monte Carlo Dropout \citep{gal2016dropout} or voxel-wise entropy, prioritize cases where the model is least confident \citep{gal2016dropout, yang2017suggestive}. Conversely, diversity-based approaches, often using core-set algorithms \citep{sener2018active} or latent space clustering, select samples that are structurally or contextually distinct from the current labeled pool to prevent redundant annotations \citep{sener2018active, smailagic2018medal}. Some frameworks attempt to combine these criteria using manually tuned weighted sums \citep{huang2010active}, adversarial representation learning \citep{sinha2019variational}, or deeply supervised strong/weak labeler schemes \citep{zhao2021dsal}. In our study, we propose to focus on a third axis: the selection should also prioritize cases where the model is most likely to produce topological errors. To this end, clDice \citep{shit2021cldice} introduces a topology-preserving loss for tubular structures, while \citet{stucki2023topologically} enforce topological faithfulness via induced matching of persistence barcodes. Motivated by the design of these topological priors, RegAL uses a topological consistency signal as a sample selection criterion, directing annotation effort toward cases that are likely to yield structurally implausible predictions. Unlike clDice and \citet{stucki2023topologically}, which inject topological priors directly into the training loss, RegAL applies this signal at the acquisition stage, complementary to any choice of segmentation loss.
%However, relying on single global heuristics like entropy often fails to capture localized topological inconsistencies (structures split into disconnected components). While effective at identifying global model confusion (e.g., severe artifacts or poor quality scans), uncertainty concentrated along small anatomical boundaries or minor structures is diluted when averaged across the entire volume, causing critical topological errors to remain underrepresented in the selection process.

\subsection{Semi-Supervised Learning in Medical Image Segmentation}
SSL is largely driven by consistency regularization and pseudo-labeling paradigms \citep{cheplygina2019not, luo2021semi}. Pseudo-labeling methods generate artificial labels for unlabeled data using model predictions, which are then treated as ground truth for subsequent training \citep{sohn2020fixmatch}. Foundational consistency methods, such as the Mean Teacher architecture \citep{tarvainen2017mean}, enforce alignment between a Student network and an exponentially moving averaged Teacher under varied perturbations \citep{tarvainen2017mean, yu2019uncertainty}.
Recent advancements introduce increasingly complex augmentation and distillation strategies to enforce this consistency \citep{bortsova2019semi, ouali2020semi}. Notable approaches include Bidirectional Copy-Paste (BCP) \citep{bai2023bidirectional}, which enforces invariance between original and cut-mixed volumes; CrossMatch (CM) \citep{zhao2024crossmatch}, which uses complex perturbation strategies combined with knowledge distillation; and MagicNet \citep{chen2023magicnet}, which employs a multi-organ partition-and-recovery strategy. Other distinct paradigms use cross-view mutual learning (CML) \citep{wu2024cross} or dynamic uncertainty-aware contrastive learning (DyCON) \citep{assefa2025dycon}. Despite their methodological sophistication, these SSL strategies share a critical vulnerability: in ultra-low data regimes (e.g., $\le 25$ samples), the initial model predictions are too inaccurate and result in confirmation bias, causing the training to collapse entirely, particularly on minor or ambiguous structures.

\subsection{Active Semi-Supervised Learning}
A natural response to annotation scarcity is to combine AL and SSL so that the same unlabeled pool both guides acquisition and supplies an auxiliary training signal. Early work in medical imaging coupled the two heuristically, e.g., for Crohn's disease segmentation \citep{mahapatra2013semi}. More recent ASSL frameworks include S4AL \citep{rangnekar2023semantic}, which interleaves uncertainty-based acquisition with consistency training, and AS3L \citep{wen2025active}, which bootstraps semi-supervised models using label propagation over self-supervised features to generate prior pseudo-labels. However, these methods stack AL and SSL as separate modules governed by disparate heuristics: the acquisition criterion and the unlabeled-data utilization objective are designed independently, producing an objective mismatch that is especially damaging in the ultra-low labeled regime. A related but distinct line of work exploits intra-volume spatial continuity: STAR \citep{ma2025spatial} selects individual 2D slices for annotation and propagates pseudo-labels to neighboring slices within the same 3D volume using spatial cross-attention, operating entirely within single-patient volumes. This intra-patient, slice-level paradigm assumes access to partially labeled volumes and is therefore fundamentally different from the inter-patient, volume-level cold-start problem addressed by RegAL, where the model must segment an entirely unseen 3D volume with no within-volume initialization. In contrast to all of the above, RegAL unifies acquisition and unlabeled-data utilization under a single topology-aware Pareto criterion applied bidirectionally, eliminating the objective mismatch that destabilizes prior ASSL pipelines.

\section{Methodology}
\textbf{Problem Formulation.} We aim to synergize AL and SSL for medical image segmentation tasks where labeled data is scarce. Let $\Omega \subset \mathbb{R}^3$ denote the discrete spatial image domain. Let $\mathcal{D}_L$ denote a small set of labeled volumes and their corresponding ground truth segmentation maps defined on $\Omega$, and $\mathcal{D}_U$ denote a much larger pool of unlabeled volumes. 
We aim to learn a network $\Psi$ to map an input volume $x$ to a dense probability map $p$ for segmenting a set of anatomical classes. We propose \textbf{RegAL}, a unified framework that tightly couples AL with SSL. At its core, RegAL is driven by a unified \textbf{Topology-Aware Pareto Optimization} that assesses images across three axes: uncertainty ($\mathcal{H}$), topological consistency ($\mathcal{T}$), and feature diversity ($\delta$). 
%We emphasize that what is shared across AL and SSL is not a single optimization objective over a unified entity set, but rather a common descriptor space and a common selection principle. Both stages operate on the same three metrics (i.e., $\mathcal{H}, \delta$, and $\mathcal{T}$) extracted by the same Teacher network, and both apply NSGA-II ranking to identify non-dominated trade-offs. What differs is the entity being ranked, unlabeled samples in AL, labeled atlases in SSL, and the direction of preference, edge cases for AL and stable anchors for SSL. 
We use the same Pareto optimization over these three axes to select unlabeled samples in AL and labeled atlases in SSL. This bidirectional reuse of a single principled mechanism is what distinguishes RegAL from prior ASSL frameworks that stack AL and SSL as separate modules with disparate heuristics.
The pipeline operates as an iterative cycle that applies this exact same multi-objective paradigm bidirectionally (Fig. \ref{fig:fig1}):
\begin{enumerate}
\item \textbf{AL via Anatomical Edge-Case Selection (Fig. \ref{fig:fig1}A):} At the start of each cycle, we apply the current segmentation model trained on the labeled set $\mathcal{D}_L$ to the unlabeled pool $\mathcal{D}_U$. We employ our multi-objective Pareto optimization to extract the first non-dominated front ($\mathcal{F}_1$) to identify the most anatomically challenging and informative "edge cases" in $\mathcal{D}_U$, which are then sent for human annotation and added to $\mathcal{D}_L$.
\item \textbf{SSL via Stable Anatomical Augmentation (Fig. \ref{fig:fig1}B):} The updated $\mathcal{D}_L$ and remaining $\mathcal{D}_U$ are then used to update the network $\Psi$ via a Mean Teacher consistency framework \citep{tarvainen2017mean}. To address the scarcity of labeled data, the same Pareto principle is repurposed inside the training loop to guide geometry-aware data augmentation. For each unlabeled sample, we extract the last Pareto front ($\mathcal{F}_{last}$) over the labeled set $\mathcal{D}_L$ to identify the most geometrically "stable" atlas. A diffeomorphic registration module then warps this atlas to the unlabeled target, creating high-fidelity, augmented training pairs for training the teacher-student segmentation model.
\end{enumerate}

\subsection{Topology-Aware AL via Pareto Optimization}
 
At the start of each active learning cycle, we use the fixed Teacher network from the previous cycle to evaluate every sample $x_u \in \mathcal{D}_U$ in the unlabeled pool. From the forward pass of each sample, we extract dense probability maps and multi-scale latent representations, which collectively serve as the basis for our selection metrics. We aim to identify the most highly uncertain, geometrically distinct, and topologically challenging "edge cases" for human annotation.  Therefore, rather than relying on a single heuristic, we formulate a multi-objective Pareto optimization focusing on the following three complementary axes:

%\paragraph{Voxel-wise Uncertainty ($\mathcal{H}$):}
%To identify samples that the model is most uncertain about the prediction, we estimate global prediction confidence via the standard mean voxel-wise entropy \citep{gal2016dropout, settles2009active}. Let $p_{c}(v)$ be the predicted probability for class $c$ at voxel $v$ within the image domain $\Omega$. The uncertainty score $\mathcal{H}(x)$ is computed as:
%\begin{equation}
%    \mathcal{H}(x) = \frac{1}{|\Omega|} \sum_{v \in \Omega} \sum_{c} -p_c(v) \log p_c(v)
%\end{equation}

\textbf{Voxel-wise Uncertainty ($\mathcal{H}$):}
For a given unlabeled image $x_u \in \mathcal{D}_U$, we estimate the global uncertainty score $\mathcal{H}(x_u)$ using the standard mean voxel-wise entropy \citep{gal2016dropout, settles2009active}. Let $p_{c}(v)$ be the predicted probability for class $c$ at voxel $v$ within the image domain $\Omega$. The uncertainty score $\mathcal{H}(x_u)$ is used to select the most uncertain cases among $\mathcal{D}_U$:
\begin{equation}
    \mathcal{H}(x_u) = \frac{1}{|\Omega|} \sum_{v \in \Omega} \sum_{c} -p_c(v) \log p_c(v)
\end{equation}

%While effective at identifying global model confusion (e.g., severe artifacts or poor quality scans), this standard metric is known to suffer from signal dilution when segmenting small anatomical targets. The high uncertainty at the structural boundaries is mathematically overpowered and averaged out by the low uncertainty of the large, confident background.

%\paragraph{Multi-Scale Feature Diversity ($\delta$):}
%In addition to selecting uncertain samples, we prevent redundant annotations by prioritizing unlabeled images that are structurally distinct from the current training set. Adapting established core-set AL principles \citep{sener2018active}, we define a diversity metric based on latent feature distance. We use a dedicated projection head on the Teacher network $\Psi$ to extract a compact, multi-scale descriptor. Let $h^{(d)}(x)$ denote the feature map output from the $d$-th downsampling stage of the encoder. We apply global average pooling (GAP) to each scale and concatenate them:
%\begin{equation}
 %   z(x) = \text{Concat}\left[ \text{GAP}(h^{(1)}(x)), \dots, \text{GAP}(h^{(4)}(x)) \right]
%\end{equation}
%For a given unlabeled image $x_u \in \mathcal{D}_U$, its diversity score $\delta(x_u)$ is computed as the minimum Euclidean distance to its nearest labeled neighbor in this multi-scale feature space: $\delta(x_u) = \min_{x_l \in \mathcal{D}_L} \| z(x_u) - z(x_l) \|_2$.

\textbf{Multi-Scale Feature Diversity ($\delta$):}
In addition to selecting uncertain samples, we prevent redundant annotations by prioritizing unlabeled images that are structurally distinct from the current training set. Adapting established core-set AL principles \citep{sener2018active}, we define a diversity metric based on latent feature distance. We use the Teacher network $\Psi$ to extract a compact, multi-scale descriptor $z$ by concatenating the global average pooled outputs from the four encoder downsampling stages. The diversity score $\delta(x_u)$ is computed as the minimum Euclidean distance to its nearest labeled neighbor in this feature space:
\begin{equation}
\delta(x_u) = \min_{x_l \in \mathcal{D}_L} \| z(x_u) - z(x_l) \|_2
\end{equation}

\textbf{Topological Consistency ($\mathcal{T}$):}
Lastly, we propose to also prioritize selecting unlabeled samples whose predicted segmentations exhibit anatomically implausible topological errors. We focus on one of the most common failure modes in low-data regimes: \emph{structural fragmentation}, where an anatomical structure that is known \emph{a priori} to form a single connected component (e.g., the brainstem or prostate transition zone) is erroneously predicted as multiple disconnected components. %Although such predictions may exhibit low local uncertainty, they represent anatomically invalid segmentations.
To identify cases with this failure, we introduce a topological consistency metric, defined on the 0-dimensional topology of the prediction (its connected components, i.e.\ the zeroth Betti number $\beta_0$), that quantifies connectedness violations.
%Because many anatomical structures are known a priori to form a single connected component, enforcing 0-dimensional topological consistency injects an anatomical prior into acquisition; this is what makes the objective anatomy-informed.
To determine which structures should satisfy this connectedness constraint, we define a subset of anatomical classes $\mathcal{C}_{conn} \subset \mathcal{C}$ that are biologically expected to form a single connected component, while excluding classes that are anatomically disjoint by definition (e.g., left and right hippocampi). Because $\mathcal{D}_L$ grows over the AL process, $\mathcal{C}_{conn}$ is determined only from the currently labeled set $\mathcal{D}_L$, never from unlabeled or held-out data: a class is included whenever it forms a single connected component in every labeled volume seen so far. This uses no information beyond what has already been manually annotated at that point in training, so it introduces no leakage from future labels.

%during preprocessing: for each class we count its connected components across all training volumes and include the class in $\mathcal{C}_{conn}$ whenever it forms a single connected component in at least 95\% of subjects. This data-driven rule requires no expert annotation and transfers across datasets.

For each class $c \in \mathcal{C}_{conn}$, we compute all connected components in the predicted mask. Let $V_{max}^{(c)}(x_u)$ denote the volume of the largest component, and $V_k^{(c)}(x_u)$ denote the volumes of all other components. %We define the fragmentation ratio for class $c$ as
%    $\sum_{k \neq \max} \frac{V_k^{(c)}}{V_{max}^{(c)}}$. 
%To amplify the penalty for severe structural fragmentation during Pareto optimization, we apply a scaling multiplier $\lambda$. 
The overall topological consistency score is the average fragmentation ratio across all relevant classes:
\begin{equation}
    \mathcal{T}(x_u) = 1 + \frac{\lambda_{conn}}{|\mathcal{C}_{conn}|} \sum_{c \in \mathcal{C}_{conn}} \sum_{k \neq \max} \frac{V_k^{(c)}(x_u)}{V_{max}^{(c)}(x_u)+\epsilon}
\end{equation}
This formulation ensures that $\mathcal{T}(x_u) = 1$ when all relevant classes form a single connected component, and $\mathcal{T}(x_u) > 1$ when fragmentation occurs, with larger values indicating more severe violations. The weight $\lambda_{conn}$ scales this objective relative to $\mathcal{H}$ and $\delta$; we fix it to $2$ in all experiments and defer its (negligible) sensitivity analysis to Section~\ref{subsec:implementation}.

%In the event that the model fails to predict any voxels for a given class, resulting in $V_{max}^{(c)} = 0$, the fragmentation ratio for that class is explicitly defined as 0, reflecting an absence of topological fragmentation.
%Larger values indicate more severe topological violations. 
%Unlike entropy-based uncertainty, this metric directly targets anatomical discontinuity. 
%While it primarily detects fragmentation (false separations) rather than erroneous fusions, fusion errors are naturally captured by the complementary uncertainty and diversity objectives within the Pareto framework.

\textbf{Active Annotation Ranking (NSGA-II Strategy):}
Lastly, we aim to select samples with simultaneously high $\mathcal{H}$, $\delta$ and $\mathcal{T}$ values. However, combining these three disparately scaled, non-commensurate metrics into a single weighted score would require fragile, manually tuned weights. We instead adopt a parameter-free formulation based on Pareto multi-objective optimization, which identifies non-dominated trade-offs among uncertainty, diversity, and topology without any scalarization. Having computed the three scalar metrics for every unlabeled sample $x_u$, we cast sample selection as simultaneously maximizing the three metrics.

To solve this, we employ the fast non-dominated sorting strategy from the NSGA-II \citep{deb2002fast} algorithm to rank all unlabeled samples into a hierarchy of Pareto fronts $\mathcal{F}_1, \mathcal{F}_2, \dots, \mathcal{F}_k$. Crucially, to distinguish between solutions within the same non-dominated front, we compute the Crowding Distance $d_i$ for every sample:
\begin{equation}
    d_i = \sum_{j=1}^{3} \frac{f_j(i+1) - f_j(i-1)}{f_j^{\max} - f_j^{\min}}
\end{equation}
where $f_j$ represents the value of the $j$-th objective function (i.e., $\mathcal{H}, \delta$, or $\mathcal{T}$). The terms $f_j(i+1)$ and $f_j(i-1)$ denote the objective values of the immediate neighboring samples when sorted along the $j$-th objective, while $f_j^{\max}$ and $f_j^{\min}$ are the maximum and minimum values of that objective within the current front. This creates a strict, global ranking where samples are prioritized first by their front index (lower is better) and second by their Crowding Distance (higher is better). 
Lastly, we select the top $N$ candidates from this globally ordered list (drawing from subsequent fronts $\mathcal{F}_2, \mathcal{F}_3, \dots$ in order whenever the first front contains fewer than $N$ samples) and send them for human annotation. %This ensures that even within equally "optimal" fronts, we explicitly prioritize samples from less dense regions of the objective space to maximize the information gain of the batch.
For AL, we prioritize samples on the first Pareto front F1 with high crowding distance to maximize diversity among selected edge cases.

\subsection{SSL with Registration-Guided Augmentation}\label{subsec:ssl_registration}

With the updated $\mathcal{D}_L$ and $\mathcal{D}_U$, we update the segmentation model using a Mean Teacher consistency framework, consisting of a Student network $\Psi'$ and a Teacher network $\Psi$ updated by exponential moving average (EMA). To handle the ultra-low sample size of $\mathcal{D}_L$ in the beginning phase, we dynamically select a reliable labeled sample (an atlas) in $\mathcal{D}_L$ that can be best registered to each unlabeled target in $\mathcal{D}_U$.  Then we perform registration-guided augmentation to produce geometrically consistent image–label training pairs for the Mean Teacher consistency framework (Fig. \ref{fig:fig1}B). %Standard data augmentation techniques (e.g., random rotation, scaling) apply arbitrary spatial transformations that fail to capture realistic anatomical shape geometries. To synthesize anatomically plausible structural variations, we embed a dynamic, registration-based geometric augmentation step directly within the semi-supervised training loop.

\textbf{Pareto Optimization for Atlas Selection:}
Using the Teacher model from the previous epoch, we dynamically select the optimal $x_l^* \in \mathcal{D}_L$ as the source atlas for each unlabeled target image $x_u$.
Crucially, we repurpose the exact same multi-objective Pareto optimization: for each labeled $x_l$, we compute uncertainty $\mathcal{H}(x_l)$, topological consistency $\mathcal{T}(x_l)$, and feature distance $\delta(x_l,x_u)=\| z(x_u) - z(x_l) \|_2$. 
%and the NSGA-II sorting framework used in our active learning cycle. 
%However, for atlas selection, we invert the optimization goal:
%For a given target $x_u$, we perform a target-conditioned Pareto ranking over the current labeled set $\mathcal{D}_L$. To ensure geometric compatibility and reliability, we seek 
Now we select the sample that \textit{minimizes} the three costs via NSGA-II sorting. Specifically, we extract the last Pareto front, $\mathcal{F}_{last}(x_u)$, containing samples that have lowest uncertainty, are most topologically intact, and are most structurally aligned with the target $x_u$. 
Within this front, we select the labeled sample $x_l^*$ that minimizes the Crowding Distance:
$x_l^* = \arg \min_{x_l \in \mathcal{F}_{last}(x_u)} d(x_l)$.
%Minimizing the crowding distance explicitly selects the sample situated in the densest, most representative region of this optimal front. This dynamic, target-aware formulation guarantees that the selected source atlas is customized to the unlabeled image's specific feature profile while being drawn from a reliable, geometric mode of the labeled distribution. 
Ultimately, this targeted selection significantly reduces the risk of generating registration artifacts caused by warping atypical or structurally distinct samples.
%For SSL atlas selection, we prefer samples on the last Pareto front with low crowding distance to select stable, representative atlases, the inverse preference to AL, reflecting the different downstream objective.
%Within each training epoch, we perform a secondary Pareto ranking over the current labeled set $\mathcal{D}_L$. Crucially, this ranking uses feature embeddings extracted by the Teacher model from the \textit{previous training epoch}.  To ensure geometric compatibility, we use the Pareto ranking in reverse. We prioritize the last Pareto front $\mathcal{F}_{last}$, which contains the "core" distribution of stable samples. Within this front, we seek the most representative anatomy by minimizing the crowding distance ($d$).  For a given target $x_u$, we deterministically select the lowest-$d$ labeled sample $x_s^*$ and its corresponding annotation $y_s$ as the optimal source atlas: $x_s^* = \arg \min_{x_l \in \mathcal{F}_{last}} d(x_l)$. This ensures the atlas represents the geometric mode of the dataset, significantly reducing the risk of artifacts caused by warping atypical geometries.

\textbf{Diffeomorphic Registration for Data Augmentation:}
We align the selected atlas $x_l^*$ to the unlabeled target $x_u$ using a VoxelMorph-based network $R_\phi$. Intensity normalization and center-of-mass alignment are first performed on $x_l^*$ and $x_u$. Then, to ensure topology preservation, we employ a diffeomorphic transformation model where the deformation field is computed by integrating a stationary velocity field $v$. The network is trained by minimizing a compound loss function:
\begin{equation}
    \mathcal{L}_{reg} = \mathcal{L}_{sim}(x_u, x_l^* \circ \phi) + \lambda_{smooth} \|\nabla v\|^2 + \lambda_{jac} \mathcal{L}_{det}(\phi)
\end{equation}
The similarity term $\mathcal{L}_{sim}$ combines Normalized Cross-Correlation and Mean Squared Error to capture both local contrast and intensity consistency. 
Standard VoxelMorph does not enforce strict topology preservation, which is critical in pathologically complex, low-label regimes where deformation fields can produce numerical folding artifacts. We therefore augment the registration objective with a Jacobian determinant penalty $L_{det}(\phi) = \mathbb{E}[(|1 - |J_{\phi}||)^2]$ following \citep{mok2020fast}, which penalizes non-diffeomorphic deformations and ensures the generated augmentation pairs remain topologically valid throughout training.
To prevent early instabilities in the registration network from generating corrupted labels, we initialize $R_\phi$ with weights pre-trained on the target dataset. During the SSL phase, $R_\phi$ is fine-tuned online, allowing the registration to adapt specifically to challenging samples. Finally, the warped image $\tilde{x} = x_l^* \circ \phi$ and the corresponding warped label $\tilde{y} = y_l^* \circ \phi$ are used as an augmented training data pair.

Note, the registration is only used as geometric augmentation, rather than label propagation. Because the atlas image and its label are warped by the same deformation field, the resulting pair $(\tilde{x}, \tilde{y})$ is internally consistent by construction. Even when the deformation field imperfectly matches the target anatomy, $(\tilde{x}, \tilde{y})$ remains a valid training sample. On the other hand, the warped label is never propagated onto $x_u$; that is, we avoid using $(x_u, \tilde{y})$ as a training data pair as a failed registration might corrupt the supervisory signal. 
\textbf{The Mean Teacher Consistency Framework:}
With the labeled and augmented training data, the Student network $\Psi'$ is updated via standard gradient descent within the training loop, while the Teacher network $\Psi$ is updated via an EMA at each training iteration $t$: $\Psi = \alpha \Psi_{t-1} + (1 - \alpha) \Psi'_t$. To ensure the scarce labeled data maintains a dominant gradient signal in ultra-low data regimes, we optimize the Student using a unified objective balancing supervised accuracy and geometric consistency:
\begin{equation}
    \mathcal{L}_{total} = \mathcal{L}_{focal}+\mathcal{L}_{dice} + \mathcal{L}_{cons}+\lambda_{TV}\mathcal{L}_{TV}.
\end{equation}
%where $x_{in}$ and $y_{in}$ denote the current input image-label pair. 
The objective is a combination of Focal Loss, soft Dice Loss, a Total Variation (TV) penalty, a consistency loss $\mathcal{L}_{cons} = \|\sigma(\Psi'(x_u)) - \sigma(\Psi(x_u))\|^2$ that enforces alignment between the Student and Teacher predictions on unlabeled samples, forcing the model to learn features robust to geometric deformations. %The supervised term $\mathcal{L}_{sup}$ is a weighted $\mathcal{L}_{sup}=$.
%The TV regularization weight is set to $0.03$ to provide a gentle smoothness constraint that suppresses voxel-wise noise without blurring sharp anatomical boundaries. We treat this composite loss as an integral part of the RegAL framework's training recipe, specifically designed to handle the anatomical complexities and synthesized data variations encountered in ultra-low data regimes.

Crucially, we implement this objective through an alternating optimization curriculum within each epoch. The Student first processes a stream of unlabeled data where input consists of registration-augmented pairs $(\tilde{x}, \tilde{y})$. This is immediately followed by a dedicated pass where input is drawn exclusively from the real labeled set $\mathcal{D}_L$. This structure ensures that while the model learns from synthesized anatomical variations, it is systematically anchored to verified human annotations at the end of every epoch, preventing representational drift.

The complete RegAL procedure, integrating the topology-aware AL cycle with the registration-guided SSL training loop under the shared Pareto criterion, is summarized in Algorithm~\ref{alg:regal}.

\begin{algorithm}[!t]
\caption{RegAL: Topology-Aware Pareto ASSL}
\label{alg:regal}
\begin{algorithmic}[1]
\Require Labeled set $\mathcal{D}_L$, unlabeled pool $\mathcal{D}_U$, Teacher $\Psi$, Student $\Psi'$, registration net $R_\phi$ (pre-trained on $\mathcal{D}_U$), AL budget $B$, per-cycle query size $N$
\Ensure Trained segmentation network $\Psi$
\While{$|\mathcal{D}_L| < B$}
    \State \textbf{// --- AL phase: anatomical edge-case selection ---}
    \ForAll{$x_u \in \mathcal{D}_U$}
        \State Compute $\mathcal{H}(x_u),\ \delta(x_u),\ \mathcal{T}(x_u)$ via Teacher $\Psi$
    \EndFor
    \State Rank $\mathcal{D}_U$ by NSGA-II (\emph{maximize} $\mathcal{H},\delta,\mathcal{T}$); take front $\mathcal{F}_1$, sort by crowding distance $d$ (high first)
    \State Query top-$N$ samples for annotation; move them $\mathcal{D}_U \!\to\! \mathcal{D}_L$
    \State Warm-start $\Psi'$ on new labels for $T_{\text{warm}}$ iterations
    \State \textbf{// --- SSL phase: registration-guided Mean Teacher ---}
    \For{each training iteration}
        \State Sample $x_u \in \mathcal{D}_U$
        \State Rank $\mathcal{D}_L$ by NSGA-II (\emph{minimize} $\mathcal{H},\mathcal{T},\delta(\cdot,x_u)$); take front $\mathcal{F}_{last}$
        \State $x_l^* \gets \arg\min_{x_l \in \mathcal{F}_{last}} d(x_l)$ \Comment{stable atlas}
        \State $\phi \gets R_\phi(x_l^*, x_u)$; $\ \tilde{x} \gets x_l^*\!\circ\phi,\ \tilde{y} \gets y_l^*\!\circ\phi$
        \State Update $\Psi'$ on $(\tilde{x},\tilde{y})$, then on real $\mathcal{D}_L$, with $\mathcal{L}_{total}$
        \State Update Teacher: $\Psi \gets \alpha\Psi + (1-\alpha)\Psi'$
        \State Fine-tune $R_\phi$ online with $\mathcal{L}_{reg}$
    \EndFor
\EndWhile
\end{algorithmic}
\end{algorithm}

\section{Experimental Configuration}

\subsection{Datasets and Ethics}

We evaluate RegAL on three publicly available medical imaging benchmarks chosen to span diverse modalities, pathologies, anatomical scales, and clinical contexts: pathological brain tumour segmentation (BraTS 2021), neonatal multi-tissue brain parcellation (dHCP), and multi-zone prostate delineation (ProstateX). All three datasets are released under open data-use agreements and were originally collected under institutional ethics approval as described below. No new patient data were collected for this study; all experiments were conducted on de-identified, publicly available images.

\textbf{BraTS 2021.}
The BraTS 2021 dataset \citep{menze2014multimodal} comprises 1{,}251 subjects with histopathologically confirmed glioma (both glioblastoma and lower-grade glioma), contributed by 23 institutions. Imaging was acquired on 3T and 1.5T scanners from multiple vendors (Siemens, Philips, GE); scanner-specific parameters varied across sites. All volumes were skull-stripped, co-registered to the SRI24 atlas, and resampled to 1\,mm isotropic voxel spacing by the challenge organisers \citep{menze2014multimodal}. We use the FLAIR modality for binary whole-tumour segmentation (tumour core + oedema), following \citep{assefa2025dycon}. Data were collected under institutional review board (IRB) approval at each contributing site; the BraTS challenge data are distributed under the RSNA-ASNR-MICCAI data-use agreement. We apply a fixed 625/626 train/test split. Gliomas present as a single infiltrative mass on FLAIR and is the only class of single connected components $\mathcal{C}_{conn} = \{\text{tumour}\}$.

\textbf{dHCP Neonatal Brain Dataset.}
The Developing Human Connectome Project (dHCP) dataset \citep{makropoulos2018developing} contains 451 neonatal subjects scanned at King's College London on a Philips Achieva 3T system equipped with a custom 32-channel neonatal head coil. T1-weighted images were acquired at 0.5\,mm isotropic resolution (TR/TE = 4795/8.7\,ms; inversion time 1740\,ms). Subjects span a gestational age range of approximately 24–45 weeks post-menstrual age (PMA) at the time of scan, covering preterm and term-equivalent neonates. Ethical approval was granted by the UK Health Research Authority (HRA; REC reference 14/LO/1169 and 14/LO/1040); written parental consent was obtained for all participants. The eight-class segmentation target covers CSF, Cortical Grey Matter (CGM), White Matter (WM), Ventricles (Vent), Cerebellum (Cer), Deep Grey Matter (DGM), Brainstem (Bs), and Hippocampus (Hippo). We use a fixed 235/216 train/test split. The set of topological constraints is $\mathcal{C}_{conn} = \{\text{CGM, WM, Cer, DGM, Bs}\}$.

\textbf{ProstateX Zonal Dataset.}
The ProstateX dataset with zonal annotations \citep{holmlund2024prostatezones} comprises 200 subjects with T2-weighted MRI acquired at Radboud University Medical Centre (RUNMC) on a Siemens Magnetom Trio 3T system (in-plane resolution $\approx$0.5\,mm; slice thickness 3.6\,mm). Subjects are men undergoing clinical prostate MRI examination. The segmentation task covers five prostate zones: Peripheral Zone (PZ), Central Zone (CZ), Transition Zone (TZ), Anterior Stroma (AS), and Urethra. Data collection was approved by the RUNMC Institutional Review Board; the dataset is publicly available through The Cancer Imaging Archive (TCIA) under a Creative Commons Attribution 3.0 Unported licence. We use a fixed 120/80 train/test split. TZ is the only class of single connected components ($\mathcal{C}_{conn} = \{\text{TZ}\}$).

\textbf{Preprocessing and Evaluation.}
To accommodate GPU memory while preserving anatomical context, BraTS and dHCP volumes are resampled to $96 \times 96 \times 96$ voxels, and ProstateX volumes to $128 \times 128 \times 32$ voxels (reflecting its anisotropic acquisition). All intensity values are normalised to $[0, 1]$ per subject. The test set is fixed prior to all experiments and is never accessed during AL acquisition cycles or SSL training. Performance is reported using three complementary metrics: the Dice Similarity Coefficient (Dice, $\uparrow$) for volumetric overlap, the Average Surface Distance (ASD, $\downarrow$, in mm) for mean boundary adherence, and the 95th-percentile Hausdorff Distance (HD95, $\downarrow$, in mm), which captures worst-case boundary error while remaining robust to a small number of outlier voxels.

\subsection{Experimental Design \& Evaluation Protocols}
\begin{figure}[!t]
    \centering
    \includegraphics[width=\linewidth]{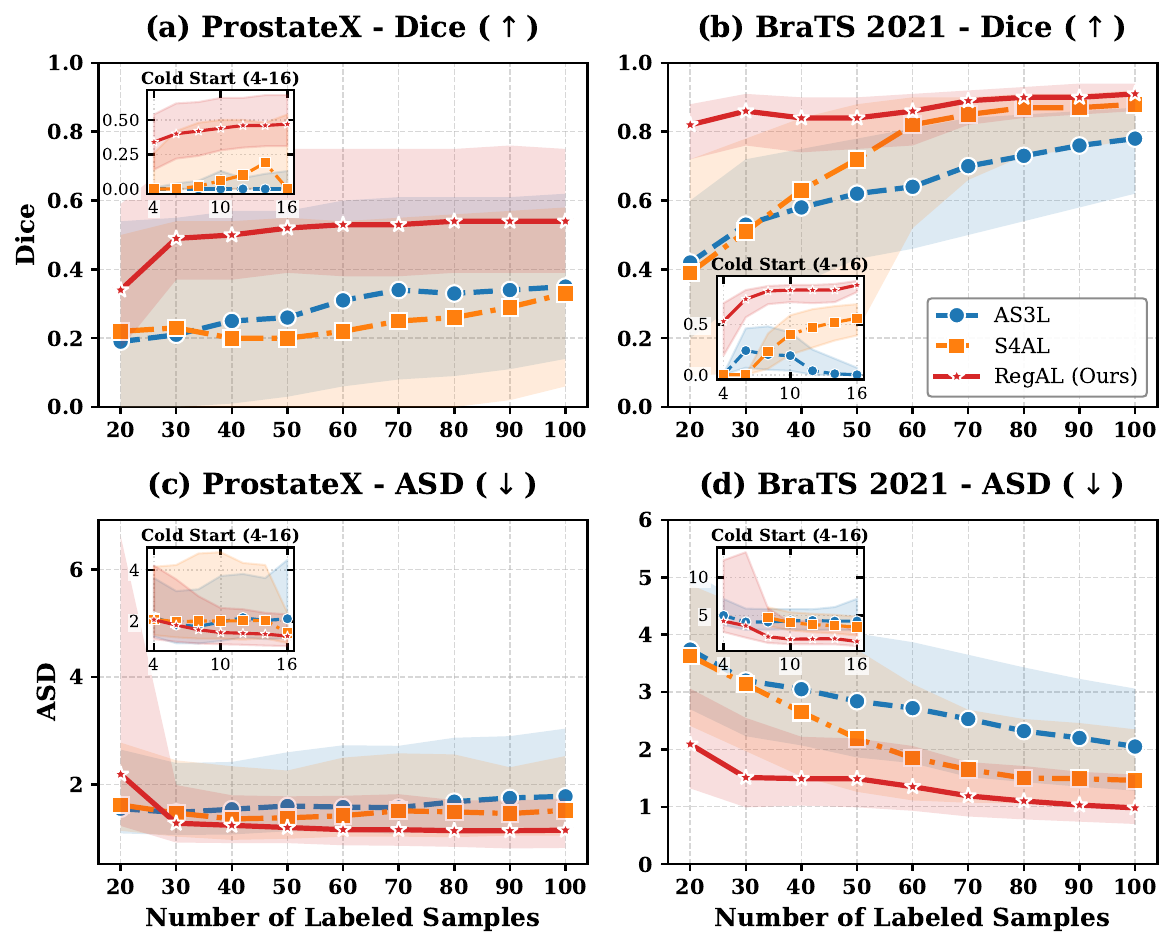}
\caption{\textbf{Segmentation accuracy for ASSL methods on ProstateX and BraTS.} Median and IQR of Dice (top row) and ASD (bottom row) against increasing annotation budgets are reported on a fixed hold-out test set. All methods are additionally evaluated on an extreme cold-start scenario starting from 4 annotated samples.}%\textbf{(a, c)} On ProstateX, standard consistency frameworks (AS3L, S4AL) suffer a severe "cold start" collapse in the low-data regime due to their inability to generate valid pseudo-labels. RegAL instantly stabilizes training, outperforming baselines across all budgets. \textbf{(b, d)} On BraTS 2021, RegAL demonstrates remarkable data efficiency, achieving high Dice and low ASD scores with only 30 samples. Shaded regions indicate the interquartile range (IQR). Detailed performance is provided in Table \ref{tab:al_comparison}.}
    \label{fig:combined_al_curves}
\end{figure}

To rigorously validate each component of our proposed RegAL framework, we design three distinct experimental protocols to evaluate the whole framework as well as the AL and SSL phase in isolation. To ensure a strictly fair comparison, every method within a protocol is evaluated under an identical fixed seed, so that all methods receive the same initial labeled pool, data ordering, and network initialisation; performance differences are therefore attributable to the method rather than to favourable random draws. Because performance under extreme scarcity is most sensitive to the choice of initial labeled pool, we separately quantify this stochastic variability by repeating the most challenging ultra-low-budget regime across five independent random initial pools on BraTS and ProstateX (Fig.~\ref{fig:combined_al_curves}) and across five seeds on the full dHCP Protocol~1 (Table~\ref{tab:dhcp_assl_comparison}), where RegAL's advantage persists in every repetition. The remaining per-budget tables report a single representative run, with Median (IQR) computed over the full held-out test set.

\textbf{Protocol 1:} 
We validate the synergistic integration of the complete RegAL framework by comparing it against two recent state-of-the-art ASSL methods: S4AL \citep{rangnekar2023semantic} and AS3L \citep{wen2025active}.
%a recent framework that bootstraps semi-supervised models by employing active learning alongside label propagation over self-supervised features to generate robust prior pseudo-labels. 
For a fair comparison, all baselines use the same 3D U-Net backbone, data splits, pre-processing, and 20{,}000-iteration training budget as RegAL, so that performance differences isolate each method's acquisition and unlabeled-data-utilisation strategy rather than implementation differences.

%This comprehensive comparison directly tests the effectiveness of our topology-aware selection and registration-guided augmentation against S4AL's uncertainty-based acquisition and AS3L's feature-driven label propagation mechanisms in ultra-low data regimes.
To evaluate performance consistency across varying anatomical complexities, we adopt a unified acquisition schedule. For both the ProstateX and BraTS datasets, we initialize with $N=20$ samples and expand to $N=100$ with a constant step size of 10 samples.
To additionally probe an \textit{extreme cold-start} regime, we repeat this comparison starting from a much smaller initial labeled pool, initializing with $N=4$ samples and expanding to $N=16$ in steps of 2, on BraTS and ProstateX across five independent random initial pools (Fig.~\ref{fig:combined_al_curves}). For dHCP, all methods already reach strong performance with 20 volumes, so we only test the extreme cold-start scenario: the labeled pool is initialized with a single volume and expanded to 3, 5, 7, and 9 volumes (Table~\ref{tab:dhcp_assl_comparison}).

\textbf{Protocol 2: } 
We evaluate the efficacy of our AL strategy independently of the SSL phase. To explicitly stress-test selection quality under \textit{extreme cold-start} conditions, we initialize the labeled pool with a single randomly selected volume before iteratively expanding to 3, 5, and 7 labeled cases. We benchmark our multi-objective Pareto optimization against Random selection (lower bound), and single-metric heuristics (uncertainty $\mathcal{H}$ \citep{gal2016dropout}, diversity $\delta$ \citep{sener2018active}). In this protocol, the same standard supervised 3D U-Net model is used for sample selection and segmentation, such that any performance gain is strictly attributable to the informativeness of the annotated data rather than the training method.

\textbf{Protocol 3:} 
To evaluate the proposed registration-guided Mean Teacher framework independently of AL, we fix the labeled set to the Pareto-selected samples by our AL method and benchmark multiple state-of-the-art SSL baselines under identical labeled data. Compared baselines represent distinct methodological paradigms: BCP \citep{bai2023bidirectional} is a consistency-based approach enforcing invariance between original and "cut-mixed" volumes. CM \citep{zhao2024crossmatch} uses complex perturbation strategies combined with knowledge distillation. CML \citep{wu2024cross} is a pseudo-labeling framework using multi-view consistency. DyCON \citep{assefa2025dycon} combines dynamic uncertainty-aware consistency with contrastive learning. MagicNet \citep{chen2023magicnet} is a partition-and-recovery strategy designed for multi-organ segmentation.

\subsection{Implementation Details}\label{subsec:implementation}

\textbf{Network Architectures:} 
We employ a modified 3D U-Net \citep{cciccek20163d} as the segmentation backbone for the Student and Teacher networks. The architecture follows a standard encoder-decoder structure with feature channels scaling from 16 to 256 (16, 32, 64, 128, 256). To ensure training stability across small batch sizes, we use Group Normalization and Leaky ReLU activations throughout the network. The registration module $R_\phi$ follows the VoxelMorph architecture \citep{balakrishnan2019voxelmorph}. It estimates a stationary velocity field at half-resolution, which is subsequently upsampled and integrated over 7 steps to generate the final diffeomorphic deformation field. To stabilize the registration network against intensity shifts, we first match the histogram of the unlabeled target $x_u$ to the source atlas $x_l^*$ before alignment. After generating the warped atlas $\tilde{x} = x_l^* \circ \phi$, we apply an inverse histogram match back to $x_u$ to ensure the final augmented sample retains the target domain's native intensity distribution.

\noindent\textbf{Training Strategy \& Optimization:} 
The framework is implemented in PyTorch and trained on NVIDIA L40S GPUs. All models are trained for a fixed duration of 20,000 iterations to ensure convergence. We optimize the network using Adam \citep{kingma2014adam} with an initial learning rate of $10^{-4}$. For the Mean Teacher consistency, the Teacher network weights $\theta'$ are updated via EMA with a decay rate of $\alpha=0.999$. We empirically set $\lambda_{TV} = 0.03$ and $\lambda_{conn} = 2$ based on a grid search in the 100-labeled setting of Protocol 3 in the ProstateX datasets (Table \ref{tab:lambda_sweep}) and apply the chosen hyperparameters to all other experiments and datasets without re-tuning. The loss components $\mathcal{L}_{focal}$, $\mathcal{L}_{dice}$, and $\mathcal{L}_{cons}$ are combined with unit weights (no additional tuning required). The consistency loss $\mathcal{L}_{cons}$ is linearly ramped up from 0 to its full contribution over the first 2{,}000 iterations to allow the Student to develop stable initial predictions before enforcing Teacher–Student agreement.

Immediately after each acquisition cycle, we run a warm-start burst of 200 iterations on the newly annotated samples before resuming the standard consistency-driven loop. Because these samples are by construction the most anatomically challenging in the pool, the burst lets the model assimilate them promptly and prevents it from treating the new labels as noise during the subsequent SSL phase.
The registration network $R_\phi$ is initialized with weights pre-trained on the target dataset using a standard unsupervised VoxelMorph objective, then fine-tuned online with a learning rate of $10^{-4}$ and hyperparameters following \citep{balakrishnan2019voxelmorph}.
Crucially, this unsupervised pre-training uses only the unlabeled training pool $\mathcal{U}$ and requires no segmentation labels, introducing no data leakage from the held-out test set. %The registration network learns only the geometric deformation statistics of the training distribution, which is standard practice in learning-based registration and independent of the downstream segmentation task.
In the registration objective, we set the smoothness penalty $\lambda_{smooth}=0.1$, following \citep{balakrishnan2019voxelmorph}, and the Jacobian determinant penalty $\lambda_{jac}=0.1$, following \citep{mok2020fast}. Supplementary Table~A1 consolidates all model, training, and hyperparameter configurations for reproducibility.

\noindent\textbf{{Compute and runtime.}}
Table~\ref{tab:efficiency} reports the computational profile of RegAL on ProstateX. Registration augmentation and Pareto selection add modest overhead per iteration, while a full 20k-iteration training run completes in approximately 4 hours on a single NVIDIA L40S with a peak VRAM footprint of 3.6\,GB, well within the capacity of a single clinical-grade GPU.

\begin{table}[!h]
\centering
\caption{Computational profiling of RegAL on ProstateX (NVIDIA L40S, single GPU).}
\label{tab:efficiency}
\begin{tabular}{lc}
\toprule
\textbf{Metric} & \textbf{Value (Mean $\pm$ Std [Range])} \\
\midrule
Registration overhead & $0.087 \pm 0.111$\,s $[0.052, 1.076]$ per pair \\
Core Pareto front calculation & $0.001 \pm 0.000$\,s per pair \\
Pareto selection latency & $0.051 \pm 0.058$\,s $[0.015, 0.367]$ per batch \\
Peak GPU VRAM & 3563\,MiB ($\approx$3.6\,GB) \\
Total wall-time (20k iterations) & $\approx$4.0\,h \\
\bottomrule
\end{tabular}
\end{table}

\subsection{Statistical Analysis}

All pairwise comparisons between RegAL and each baseline are conducted using the two-sided Wilcoxon signed-rank test \citep{wilcoxon1945individual}, a non-parametric paired test appropriate for the skewed, bounded distributions of Dice, ASD, and HD95 scores. The test is applied per-subject on held-out test sets ($n=626$ for BraTS 2021, $n=216$ for dHCP, $n=80$ for ProstateX), treating each subject as a matched pair across methods. To control the family-wise error rate across multiple baseline comparisons within each experimental condition (dataset $\times$ annotation budget $\times$ metric), $p$-values are adjusted using the Holm--Bonferroni procedure \citep{holm1979simple}. 
%Results are reported in tables using superscripts: $^{*}p{<}0.05$, $^{**}p{<}0.01$, $^{***}p{<}0.001$; cells without a superscript are not significantly different from RegAL. Bold formatting indicates the best numerical value; underline indicates the second-best. Where a baseline achieves a numerically superior metric value (e.g., ASD at very low annotation budgets on ProstateX), the significance marker reflects statistical difference in either direction, with bold and underline formatting indicating which method is numerically better. 
Except where multiple seeds are explicitly reported (the five-seed extreme cold-start experiment and the dHCP Protocol~1 comparison), these tests are conditional on a single training run per method: the reported $p$-values characterise per-subject differences between the specific models compared rather than method-level variance across training runs, and should be interpreted together with the cross-seed stability evidence. %Given the large test-set sizes, we therefore emphasise effect magnitudes (the tabulated metric differences) alongside the significance markers.

\section{Results and Discussion}

To demonstrate the generalization ability and anatomical robustness of RegAL, we conduct extensive validation across three distinct medical imaging benchmarks representing diverse modalities, anatomical structures, and pathological complexities: the ProstateX dataset for multi-class prostate zone segmentation, the BraTS dataset for heterogeneous brain tumor segmentation, and the dHCP (Developing Human Connectome Project) dataset for preterm infant brain tissue segmentation.

\subsection{Protocol 1: Comparing with ASSL Baselines}
\textit{Protocol 1} reveals a stark contrast in learning efficiency (Fig.~\ref{fig:combined_al_curves}).
On ProstateX at the 30-sample starting point, RegAL achieves a Median Dice of 0.49 (IQR 0.35), more than doubling AS3L (0.21) and S4AL (0.23).
On BraTS, RegAL reaches a Dice of 0.86 (0.15) with only 30 samples, approaching fully-supervised performance significantly faster than competing methods.
The advantage extends to boundary adherence: on BraTS, RegAL achieves an ASD of 1.51 (1.56)\,mm, effectively halving the error of AS3L (3.20\,mm) and S4AL (3.14\,mm); on ProstateX, RegAL maintains the lowest ASD of 1.28 (1.07)\,mm across all budgets.
Both baselines exhibit severe performance stagnation at budgets $\leq 50$ samples, whereas RegAL instantly stabilises.
Per-region breakdowns are reported in Supplementary Table~A2.

Motivated by this large performance gap, we further examine a more challenging \textit{extreme cold-start} scenario in which the initial labeled pool contains only 4 samples and expands to 16 in steps of 2. To make the comparison direct, we evaluate AS3L and S4AL under the identical schedule (Fig.~\ref{fig:combined_al_curves}). Both baselines remain unstable in this regime, whereas RegAL trains stably and improves rapidly from the first few labels, retaining a clear margin over both across the entire 4-to-16 budget range.  With only a handful of labels, RegAL reaches the accuracy of the baselines attain at substantially larger budgets. %This shows the strong data efficiency of the proposed framework.
%The dHCP results are reported at smaller annotation budgets than BraTS and ProstateX, and it is worth stating why. Because the neonatal parcellation task is comparatively high-contrast, RegAL already reaches strong performance with as few as 3 labeled volumes in Protocol~1 and saturates at these very low budgets; evaluating larger labeled sets would add computational cost without changing the conclusions. We therefore focus the dHCP evaluation on Protocol~1 and the ultra-low SSL regime (5 and 10 labeled samples in Protocol~3), which also best reflects the practical annotation constraints of this cohort.
This behavior is mirrored under extreme low-label constraints on the dHCP cohort (Table~\ref{tab:dhcp_assl_comparison}). The per-class breakdown in Supplementary Table~A3 shows that at a scarce budget of 9 labeled brains, S4AL achieves a Median Dice of only 0.16 on the Hippocampus, a small structure particularly vulnerable to topological collapse in cold-start conditions, whereas RegAL maintains robust performance at 0.79.

\begin{figure}[!t]
    \centering
    \includegraphics[width=\textwidth]{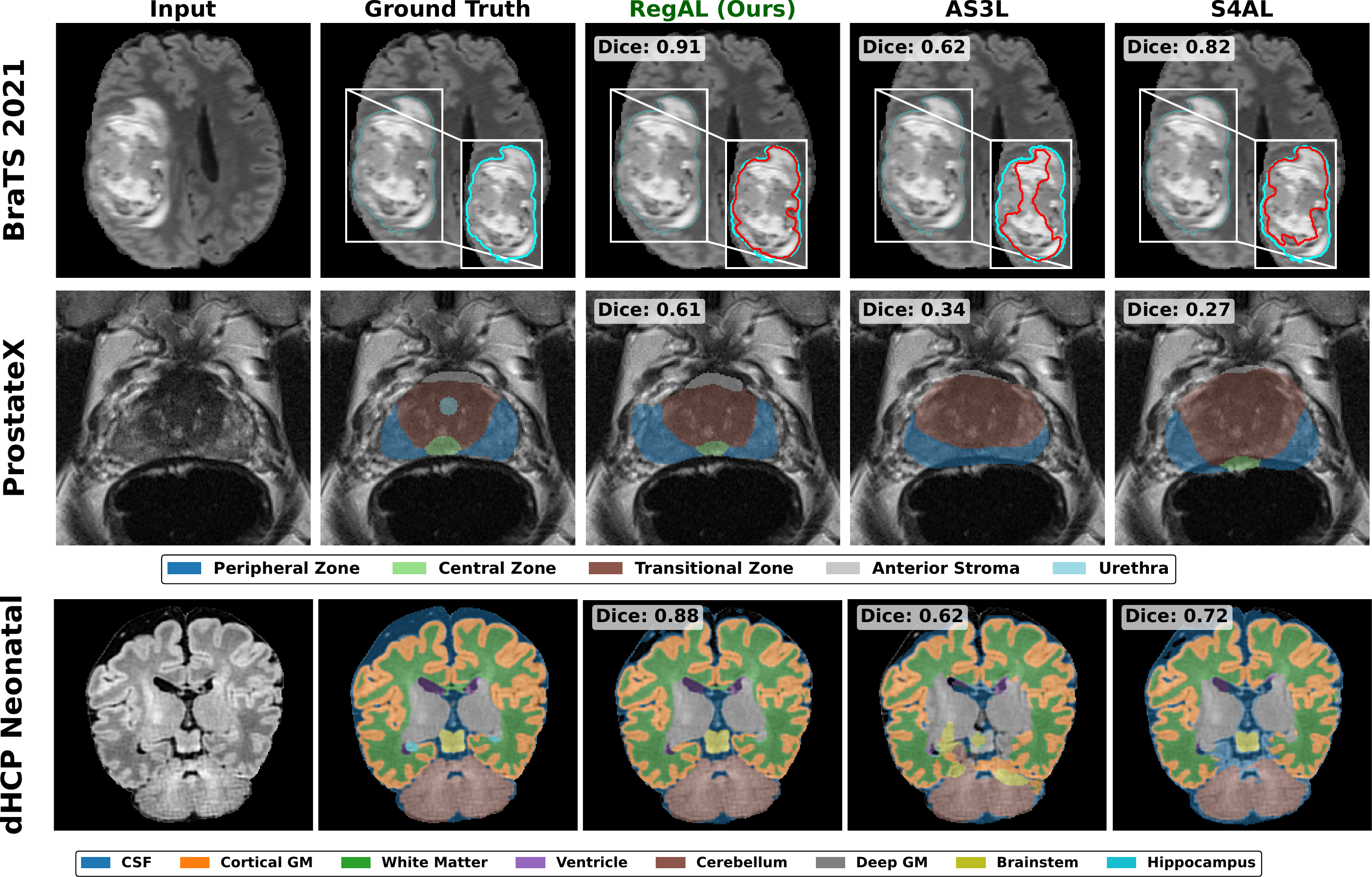}
    \caption{\textbf{Qualitative comparison of segmentation results on BraTS 2021 (top row, 50 labeled), ProstateX (middle row, 20 labeled), and dHCP (bottom row, 9 labeled).}
    Columns display (left to right): Input MRI, Ground Truth, RegAL (Ours), AS3L, and S4AL.
    \textbf{Top Row:} Red contours indicate model predictions, while cyan contours represent the Ground Truth. Zoomed insets highlight boundary precision.
    \textbf{Middle Row:} Multi-class segmentation of prostate zones (legend at bottom).
    \textbf{Bottom Row:} Eight-class neonatal brain tissue segmentation on dHCP; RegAL preserves topological coherence on subcortical structures (Ventricles, Hippocampus) where the ASSL baselines fragment.}
    \label{fig:qualitative_results}
\end{figure}

Qualitative results in Fig.~\ref{fig:qualitative_results} confirm the advantage of RegAL across all three datasets. We visualize segmentation quality at the annotation budget where RegAL's performance plateaus. With 50 labeled samples, RegAL produces smooth tumor boundaries on BraTS; with 20 labeled samples, it recovers minor prostate zones on ProstateX; and with 9 labeled samples, it preserves topological coherence on subcortical dHCP structures, whereas the baselines still frequently under-segment pathologically complex regions and yield topologically implausible results at those annotation budgets. Supplementary Fig.~A1 shows the same cases in the sagittal plane, confirming that this advantage is consistent across orthogonal views.

Taken together, these results demonstrate that RegAL successfully mitigates the cold-start problem that causes existing active semi-supervised frameworks to collapse in ultra-low data regimes. This stability follows from three design choices that share a single multi-objective Pareto criterion rather than competing as independently tuned modules: registration-guided augmentation supplies internally consistent synthetic supervision from the very first iteration (Section~\ref{subsec:ssl_registration}), the topological consistency metric $\mathcal{T}$ recovers the structural failure modes that voxel-wise uncertainty alone dilutes away (Table~\ref{tab:brats_ablation_overall}), and the bidirectional use of the Pareto front, $\mathcal{F}_1$ to expand the labeled set and $\mathcal{F}_{last}$ to anchor atlas selection, couples acquisition and augmentation under that same criterion (Section~\ref{subsec:ssl_registration}). Because all three follow from one shared criterion rather than three separate heuristics, they reinforce one another instead of working at cross purposes, which is why the cold-start collapse seen in prior ASSL baselines does not occur here.

\begin{table}[!t]
\centering
\caption{Comparison of ASSL frameworks on dHCP across annotation budgets, aggregated over all 8 tissue classes and 5 random seeds. Median (IQR); ASD and HD95 in mm. \textbf{Bold}: best; \underline{underline}: second-best. Superscripts: $^{**}p{<}0.01$, $^{***}p{<}0.001$ vs.\ RegAL (Wilcoxon, Holm--Bonferroni). %At N=1 (single atlas), registration-guided augmentation produces noisier boundaries; from N$\geq$3 RegAL leads all metrics. 
Per-class breakdown is provided in Supplementary Table~A3.}
\label{tab:dhcp_assl_comparison}
\setlength{\tabcolsep}{6pt}
\begin{tabularx}{\textwidth}{c X ccc}
\toprule
\textbf{Budget} & \textbf{Method} & \textbf{Dice} $\uparrow$ & \textbf{ASD} $\downarrow$ & \textbf{HD95} $\downarrow$ \\
\midrule
\multirow{3}{*}{\textbf{1 Labeled}}
 & AS3L \citep{wen2025active}         & \underline{0.55 (0.30)}$^{***}$ & \textbf{4.76 (4.18)}$^{***}$    & \textbf{15.59 (15.06)}$^{**}$ \\
 & S4AL \citep{rangnekar2023semantic} & 0.54 (0.30)$^{***}$             & \underline{4.85 (5.32)}$^{***}$ & 16.90 (16.63)$^{***}$ \\
 & RegAL (Ours)                       & \textbf{0.62 (0.22)}            & 5.54 (4.64)                      & \underline{16.19 (14.59)} \\
\midrule
\multirow{3}{*}{\textbf{3 Labeled}}
 & AS3L \citep{wen2025active}         & \underline{0.61 (0.29)}$^{***}$ & 3.81 (3.91)$^{***}$             & \underline{13.14 (13.20)}$^{***}$ \\
 & S4AL \citep{rangnekar2023semantic} & 0.58 (0.32)$^{***}$             & \underline{3.11 (3.90)}$^{***}$ & 14.70 (19.56)$^{***}$ \\
 & RegAL (Ours)                       & \textbf{0.84 (0.09)}            & \textbf{1.32 (0.83)}            & \textbf{4.13 (3.38)} \\
\midrule
\multirow{3}{*}{\textbf{5 Labeled}}
 & AS3L \citep{wen2025active}         & 0.63 (0.26)$^{***}$             & 4.38 (5.36)$^{***}$             & 14.26 (15.74)$^{***}$ \\
 & S4AL \citep{rangnekar2023semantic} & \underline{0.67 (0.29)}$^{***}$ & \underline{2.17 (2.56)}$^{***}$ & \underline{12.01 (16.78)}$^{***}$ \\
 & RegAL (Ours)                       & \textbf{0.85 (0.08)}            & \textbf{1.30 (0.95)}            & \textbf{3.80 (4.07)} \\
\midrule
\multirow{3}{*}{\textbf{7 Labeled}}
 & AS3L \citep{wen2025active}         & 0.64 (0.25)$^{***}$             & 4.35 (4.73)$^{***}$             & 12.86 (11.91)$^{***}$ \\
 & S4AL \citep{rangnekar2023semantic} & \underline{0.70 (0.26)}$^{***}$ & \underline{2.02 (2.98)}$^{***}$ & \underline{10.85 (15.40)}$^{***}$ \\
 & RegAL (Ours)                       & \textbf{0.87 (0.06)}            & \textbf{1.08 (0.64)}            & \textbf{3.03 (2.47)} \\
\midrule
\multirow{3}{*}{\textbf{9 Labeled}}
 & AS3L \citep{wen2025active}         & 0.65 (0.24)$^{***}$             & 4.26 (4.43)$^{***}$             & 12.55 (11.59)$^{***}$ \\
 & S4AL \citep{rangnekar2023semantic} & \underline{0.73 (0.24)}$^{***}$ & \underline{1.82 (2.35)}$^{***}$ & \underline{9.04 (12.36)}$^{***}$ \\
 & RegAL (Ours)                       & \textbf{0.87 (0.06)}            & \textbf{1.05 (0.62)}            & \textbf{2.86 (2.05)} \\
\bottomrule
\end{tabularx}
\end{table}
%Similarly, on BraTS (Fig.~\ref{fig:combined_al_curves}b), RegAL reaches a 0.86 (0.15) Dice score with only 30 samples, approaching the fully supervised oracle ceiling of 0.90 significantly faster than competing methods. Crucially, this performance advantage extends to boundary adherence (Fig.~\ref{fig:combined_al_curves}c, d). On the structurally complex BraTS task, RegAL reduces the ASD to 1.51 (1.56) mm at the 30-sample starting point, effectively halving the error rates of AS3L (3.20 mm) and S4AL (3.14 mm). A similar trend is observed on ProstateX, where RegAL maintains the lowest ASD of 1.28 (1.07) mm, indicating that our registration-guided augmentation effectively preserves topological integrity where standard pseudo-labeling methods often generate fragmented predictions. 

\subsection{Protocol 2: Ablation on AL Selection Strategy}
In \textit{Protocol 2}, we stress-test the AL selection strategy in isolation (a plain supervised U-Net, no SSL) under extreme cold-start. On the binary BraTS tumour segmentation task, the Pareto objective achieves the best performance across all reported metrics (Dice and ASD) at 3-, 5-, and 7-label budgets, substantially outperforming single-metric heuristics (Table~\ref{tab:brats_ablation_overall}); notably, it attains the lowest ASD values in all regimes, indicating improved boundary adherence and reduced structural fragmentation, precisely the failure mode targeted by our topology-aware objective. Uncertainty alone underperforms here because a 3D volume is mostly background that the model already segments with high confidence; averaging voxel-wise entropy over the whole volume lets this large, confident region drown out the much smaller signal at structural boundaries. The topological consistency metric $\mathcal{T}$ instead directly targets samples with structural disconnection, the dominant error mode in this regime, forcing the annotator to correct structural ``silent failures'' that entropy-only strategies miss. 

\begin{table}[!t]
\centering
\caption{Comparison of AL Strategies on the BraTS 2021 Dataset. We report Median (IQR) for Dice ($\uparrow$), ASD ($\downarrow$, mm), and HD95 ($\downarrow$, mm) across annotation budgets. \textbf{Bold} indicates the best performance; \underline{underline} indicates the second-best. Superscripts denote significance vs.\ Pareto($\mathcal{H}$, $\mathcal{T}$, $\delta$) (Wilcoxon, Holm--Bonferroni): $^{**}p{<}0.01$, $^{***}p{<}0.001$. HD95 not available for Random.}
\label{tab:brats_ablation_overall}
\setlength{\tabcolsep}{3pt}
\resizebox{\textwidth}{!}{%
\begin{tabular}{l ccc | ccc | ccc}
\toprule
& \multicolumn{3}{c|}{\textbf{3 Labeled}} & \multicolumn{3}{c|}{\textbf{5 Labeled}} & \multicolumn{3}{c}{\textbf{7 Labeled}} \\
\textbf{Method} & \textbf{Dice} $\uparrow$ & \textbf{ASD} $\downarrow$ & \textbf{HD95} $\downarrow$ & \textbf{Dice} $\uparrow$ & \textbf{ASD} $\downarrow$ & \textbf{HD95} $\downarrow$ & \textbf{Dice} $\uparrow$ & \textbf{ASD} $\downarrow$ & \textbf{HD95} $\downarrow$ \\
\midrule
\multicolumn{10}{l}{\textit{\textbf{Standard Baselines}}} \\
Random                                      & 0.69 (0.19)              & \underline{3.03 (2.44)} & --                     & 0.69 (0.17)              & 3.15 (2.62)             & --                     & 0.73 (0.19)              & 2.84 (1.88)             & -- \\
Uncertainty \citep{gal2016dropout} ($\mathcal{H}$) & 0.60 (0.24)$^{***}$ & 3.39 (2.56)$^{**}$  & 15.80 (13.48)$^{***}$  & \underline{0.76 (0.23)}$^{***}$ & \underline{2.52 (2.42)}$^{***}$ & \underline{10.22 (12.02)}$^{***}$ & \underline{0.79 (0.23)}$^{***}$ & \underline{2.27 (2.54)}$^{***}$ & \underline{10.30 (15.71)}$^{***}$ \\
Diversity \citep{sener2018active} ($\delta$) & \underline{0.77 (0.26)}$^{***}$ & 3.33 (4.04)$^{***}$ & \underline{11.18 (15.74)}$^{***}$ & 0.75 (0.26)$^{***}$ & 2.94 (3.46)$^{***}$ & 12.08 (12.17)$^{***}$ & 0.78 (0.23)$^{***}$ & 2.90 (3.88)$^{***}$ & 11.92 (12.81) \\
Topology $\mathcal{T}$                      & 0.26 (0.19)$^{***}$      & 3.93 (2.22)$^{***}$ & 30.48 (16.65)$^{***}$  & 0.14 (0.15)$^{***}$      & 4.83 (4.17)$^{***}$     & 32.08 (22.65)$^{***}$  & 0.37 (0.30)$^{***}$      & 3.88 (2.47)$^{***}$     & 23.79 (18.73)$^{***}$ \\
\midrule
Pareto($\mathcal{H}$, $\mathcal{T}$, $\delta$) & \textbf{0.84 (0.18)} & \textbf{2.06 (3.42)} & \textbf{7.97 (16.61)} & \textbf{0.85 (0.17)} & \textbf{1.96 (2.69)} & \textbf{8.06 (13.47)} & \textbf{0.82 (0.18)} & \textbf{1.86 (1.89)} & \textbf{9.35 (13.32)} \\
\bottomrule
\end{tabular}
}
\end{table}

\begin{table}[!t]
\centering
\caption{Comparison of AL strategies on the ProstateX dataset (Protocol 2), aggregated over all five prostate zones. We report Median (IQR) for Dice ($\uparrow$), ASD ($\downarrow$, mm), and HD95 ($\downarrow$, mm) across annotation budgets. \textbf{Bold}: best; \underline{underline}: second-best. Superscripts denote significance vs.\ Pareto($\mathcal{H}$, $\mathcal{T}$, $\delta$) (Wilcoxon, Holm--Bonferroni): $^{*}p{<}0.05$, $^{***}p{<}0.001$. The per-zone breakdown is given in Supplementary Table~A4.}
\label{tab:prostatex_summary_overall}
\setlength{\tabcolsep}{3pt}
\resizebox{\textwidth}{!}{%
\begin{tabular}{l ccc | ccc | ccc}
\toprule
& \multicolumn{3}{c|}{\textbf{3 Labeled}} & \multicolumn{3}{c|}{\textbf{5 Labeled}} & \multicolumn{3}{c}{\textbf{7 Labeled}} \\
\textbf{Method} & \textbf{Dice} $\uparrow$ & \textbf{ASD} $\downarrow$ & \textbf{HD95} $\downarrow$ & \textbf{Dice} $\uparrow$ & \textbf{ASD} $\downarrow$ & \textbf{HD95} $\downarrow$ & \textbf{Dice} $\uparrow$ & \textbf{ASD} $\downarrow$ & \textbf{HD95} $\downarrow$ \\
\midrule
Random                                       & \underline{0.19 (0.28)} & 27.65 (32.76) & 58.20 (56.56) & 0.24 (0.30)$^{***}$ & 16.34 (26.97)$^{***}$ & 43.84 (48.22)$^{***}$ & 0.27 (0.37) & 9.84 (23.32) & 34.50 (48.15) \\
Uncertainty \citep{gal2016dropout} ($\mathcal{H}$) & 0.17 (0.41)$^{*}$ & \underline{14.00 (32.76)}$^{*}$ & 40.63 (52.63)$^{*}$ & 0.19 (0.45)$^{***}$ & \textbf{3.45 (10.35)}$^{*}$ & \textbf{21.55 (29.47)} & 0.27 (0.45)$^{*}$ & \textbf{2.73 (6.29)}$^{***}$ & \underline{15.20 (18.67)}$^{***}$ \\
Diversity \citep{sener2018active} ($\delta$) & 0.00 (0.10)$^{***}$ & 66.59 (77.50)$^{***}$ & 126.82 (105.85)$^{***}$ & 0.21 (0.37)$^{***}$ & 18.31 (42.17)$^{***}$ & 44.48 (102.38)$^{***}$ & 0.28 (0.48)$^{*}$ & 6.39 (18.98) & 24.51 (49.51) \\
Topology ($\mathcal{T}$) & 0.18 (0.43) & \textbf{11.92 (26.81)} & \textbf{30.01 (56.65)}$^{*}$ & \underline{0.27 (0.50)}$^{***}$ & \underline{6.37 (23.89)}$^{*}$ & \underline{25.87 (75.65)} & \underline{0.28 (0.49)} & \underline{3.35 (7.02)}$^{*}$ & \textbf{15.09 (23.46)}$^{*}$ \\
\midrule
Pareto($\mathcal{H}$, $\mathcal{T}$, $\delta$) & \textbf{0.21 (0.40)} & 16.38 (44.19) & \underline{39.80 (97.26)} & \textbf{0.30 (0.44)} & \underline{4.43 (16.07)} & \underline{22.56 (44.76)} & \textbf{0.31 (0.44)} & 5.33 (17.94) & 21.28 (42.69) \\
\bottomrule
\end{tabular}
}
\end{table}

\begin{table}[!t]
\centering
\caption{Comparison of AL strategies on the dHCP dataset (Protocol 2), aggregated over all eight tissue classes. We report Median (IQR) for Dice ($\uparrow$), ASD ($\downarrow$, mm), and HD95 ($\downarrow$, mm) across annotation budgets. \textbf{Bold}: best; \underline{underline}: second-best. Superscripts denote significance vs.\ Pareto($\mathcal{H}$, $\mathcal{T}$, $\delta$) (per-subject macro score, two-sided Wilcoxon signed-rank, Holm--Bonferroni, $n{=}216$): $^{*}p{<}0.05$, $^{**}p{<}0.01$, $^{***}p{<}0.001$; no superscript $=$ not significant. Owing to the large sample size, paired differences can reach significance even where medians coincide (e.g., Topology vs.\ Pareto at $N{=}7$ HD95). The per-class breakdown is given in Supplementary Table~A5.}
\label{tab:dhcp_al_overall}
\setlength{\tabcolsep}{3pt}
\resizebox{\textwidth}{!}{%
\begin{tabular}{l ccc | ccc | ccc}
\toprule
& \multicolumn{3}{c|}{\textbf{3 Labeled}} & \multicolumn{3}{c|}{\textbf{5 Labeled}} & \multicolumn{3}{c}{\textbf{7 Labeled}} \\
\textbf{Method} & \textbf{Dice} $\uparrow$ & \textbf{ASD} $\downarrow$ & \textbf{HD95} $\downarrow$ & \textbf{Dice} $\uparrow$ & \textbf{ASD} $\downarrow$ & \textbf{HD95} $\downarrow$ & \textbf{Dice} $\uparrow$ & \textbf{ASD} $\downarrow$ & \textbf{HD95} $\downarrow$ \\
\midrule
Random & 0.79 (0.17)$^{***}$ & 0.75 (0.54)$^{***}$ & 2.06 (1.92)$^{***}$ & \textbf{0.83 (0.13)} & 0.55 (0.31) & \textbf{1.50 (1.01)} & \textbf{0.85 (0.13)} & \textbf{0.48 (0.24)} & \textbf{1.41 (0.75)} \\
Uncertainty \citep{gal2016dropout} ($\mathcal{H}$) & \underline{0.81 (0.15)}$^{***}$ & \underline{0.65 (0.39)}$^{***}$ & \underline{2.00 (1.20)}$^{***}$ & \underline{0.82 (0.14)}$^{***}$ & 0.59 (0.32)$^{***}$ & 1.80 (1.09)$^{***}$ & 0.83 (0.14)$^{***}$ & 0.59 (0.32)$^{***}$ & 1.66 (1.03)$^{***}$ \\
Diversity \citep{sener2018active} ($\delta$) & \underline{0.81 (0.15)}$^{***}$ & 0.70 (0.42)$^{***}$ & 2.06 (1.29)$^{***}$ & \underline{0.82 (0.14)}$^{***}$ & 0.60 (0.32)$^{***}$ & 1.80 (1.09)$^{***}$ & 0.82 (0.14)$^{***}$ & 0.62 (0.36)$^{***}$ & 1.80 (1.09)$^{***}$ \\
Topology ($\mathcal{T}$) & 0.76 (0.20)$^{***}$ & 0.76 (0.38)$^{***}$ & 2.24 (1.11)$^{***}$ & \underline{0.82 (0.15)}$^{**}$ & \underline{0.54 (0.27)} & \underline{1.58 (0.90)} & \underline{0.84 (0.14)} & \underline{0.49 (0.24)} & \underline{1.50 (0.75)}$^{**}$ \\
\midrule
Pareto($\mathcal{H}$, $\mathcal{T}$, $\delta$) & \textbf{0.82 (0.16)} & \textbf{0.60 (0.31)} & \textbf{1.80 (1.03)} & \textbf{0.83 (0.14)} & \textbf{0.53 (0.25)} & \textbf{1.50 (0.84)} & \textbf{0.85 (0.13)} & \textbf{0.48 (0.23)} & \underline{1.50 (0.75)} \\
\bottomrule
\end{tabular}
}
\end{table}

The overall advantage of the Pareto objective generally holds on the other two datasets, though the magnitude of the advantage differs. On dHCP (Table~\ref{tab:dhcp_al_overall}), the Pareto objective is best or second-best in all scenarios. Its advantage is most pronounced in the extreme cold-start regime ($N{=}3$), where it significantly outperforms every single-metric heuristic and random selection on all three metrics (Dice, ASD, and HD95; all $p{<}0.001$). As the budget increases to $N{=}5$ and $N{=}7$, performance differences become narrow and are frequently no longer significant. This pattern is consistent with dHCP's comparatively high tissue contrast, where even simpler selection strategies reach strong performance once a handful of labeled volumes are available. On the harder multi-zone ProstateX task (Table~\ref{tab:prostatex_summary_overall}), although the Pareto objective attains the highest Dice at every budget, differences in ASD and HD95 are less consistent. This variability likely reflects the frequent "near-empty'' predictions in the low-data regime disproportionately affecting boundary-distance metrics. Nevertheless, the consistent Dice advantage indicates that combining uncertainty, topology, and diversity provides a more reliable acquisition strategy than any individual criterion. The per-zone and per-class breakdowns are given in Supplementary Tables~A4 and~A5, respectively.
%  end hidden T-ablation 

%—precisely the failure mode targeted by our topology-aware objective.

% [ProstateX Protocol-2 discussion folded into the Protocol-2 intro paragraph above; aggregate table demoted, per-zone results in Supplementary Table~A4.]

\begin{figure}[htbp]
  \centering
  \small 
  \includegraphics[width=0.78\linewidth]{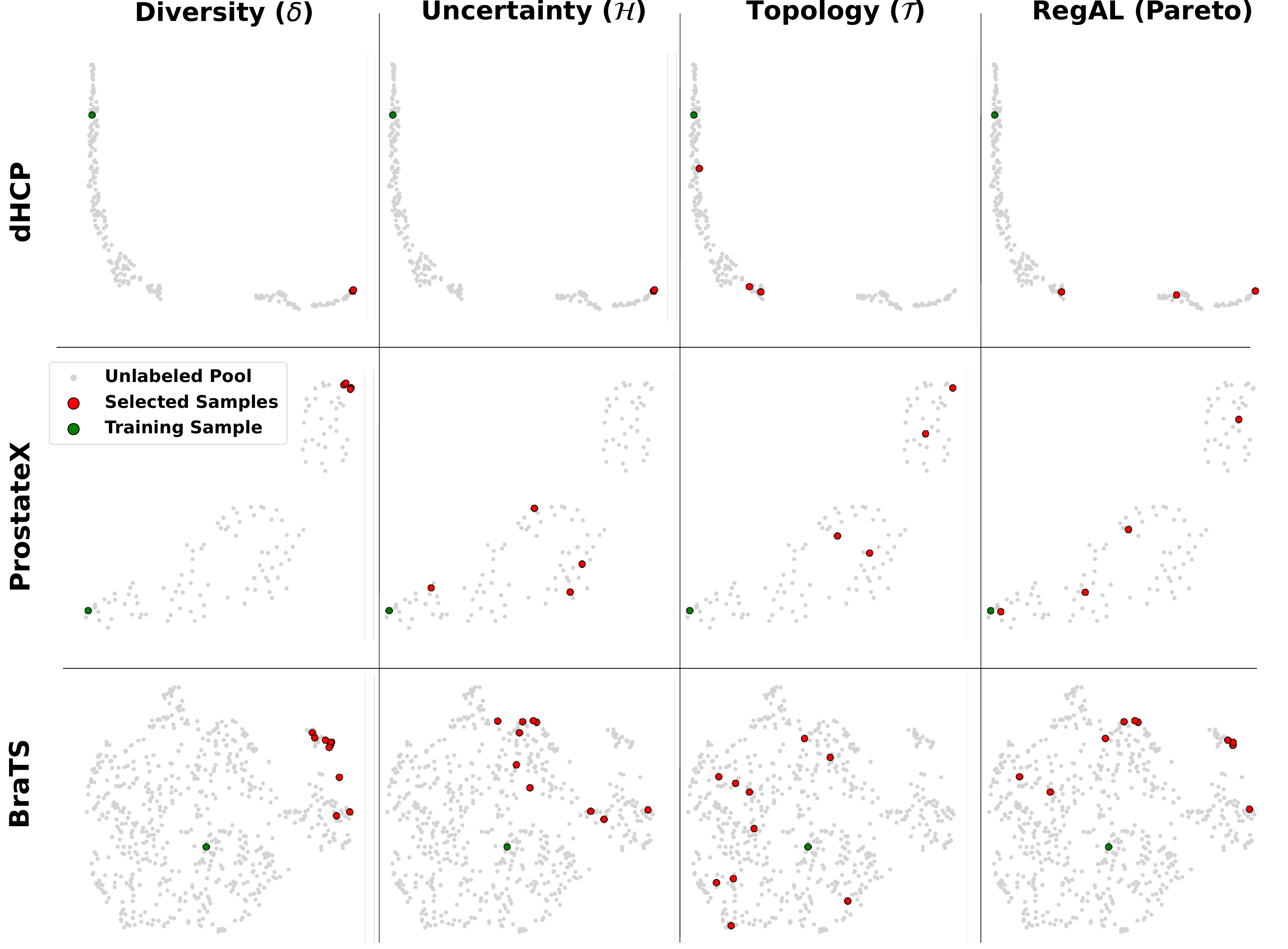}
  \caption{Visualization of active sample selection in the encoder feature space across all three datasets, starting from a single initial training sample. Each panel shows a 2D UMAP projection of the unlabeled pool (grey), the single initial training sample (green), and the samples each strategy selects next (red): 3 additional samples for dHCP, 4 for ProstateX, and 10 for BraTS, chosen for clarity. Single-metric heuristics, i.e. diversity ($\delta$), uncertainty ($\mathcal{H}$), and topology ($\mathcal{T}$), tend to cluster selected samples in a narrow region of the feature space, reflecting their single-axis bias. In contrast, RegAL consistently selects samples that are spatially distributed across the feature space, covering both high-uncertainty boundary regions and structurally distinct clusters, demonstrating the complementary coverage achieved by multi-objective Pareto optimization.}
  \label{fig:latent_space_comparison}
\end{figure}

In Fig. \ref{fig:latent_space_comparison}, we further visualize, starting from a single initial training sample, where each strategy's subsequently selected samples land in the encoder feature space via UMAP projection. To maintain visual readability given the differing spatial distributions of the UMAP embeddings, we display 3 additional samples for dHCP, 4 for ProstateX, and 10 for BraTS. While single-metric heuristics concentrate selections in narrow regions of the feature space reflecting their individual biases, the Pareto optimization of RegAL consistently distributes selected samples across structurally distinct clusters, confirming that multi-objective optimization achieves broader and more complementary coverage than any single criterion alone.

%\pagebreak
\subsection{Protocol 3: Registration-Guided SSL}
In \textit{Protocol 3}, we isolate the SSL component by fixing the labeled set and benchmarking multiple SSL strategies under identical annotation budgets (Tables \ref{tab:brats_median_iqr_updated}, \ref{tab:prostatex_summary_final}, \& ~\ref{tab:dhcp_executive_summary_no_oracle}). Segmentation accuracy in individual anatomical regions is detailed in Supplementary Tables A6 \& A7, while qualitative comparisons are provided in Supplementary Figs.\ A2 \& A3.

\begin{table}[!t]
\centering
\caption{Comparison of SSL Strategies on the BraTS 2021 Dataset. We report Median (IQR) for Dice ($\uparrow$), ASD ($\downarrow$, mm), and HD95 ($\downarrow$, mm) across annotation budgets. \textbf{Bold} indicates the best value; \underline{underline} indicates the second-best among SSL methods. Superscripts denote statistical significance vs.\ RegAL (two-sided Wilcoxon signed-rank test, Holm--Bonferroni corrected, $n{=}626$): $^{**}p{<}0.01$, $^{***}p{<}0.001$.}
\label{tab:brats_median_iqr_updated}
\setlength{\tabcolsep}{2pt}
\resizebox{\textwidth}{!}{%
\begin{tabular}{l ccc | ccc | ccc}
\toprule
& \multicolumn{3}{c|}{\textbf{10 Labeled}} & \multicolumn{3}{c|}{\textbf{20 Labeled}} & \multicolumn{3}{c}{\textbf{100 Labeled}} \\
\textbf{Method} & \textbf{Dice} $\uparrow$ & \textbf{ASD} $\downarrow$ & \textbf{HD95} $\downarrow$ & \textbf{Dice} $\uparrow$ & \textbf{ASD} $\downarrow$ & \textbf{HD95} $\downarrow$ & \textbf{Dice} $\uparrow$ & \textbf{ASD} $\downarrow$ & \textbf{HD95} $\downarrow$ \\
\midrule
U-Net (Baseline) & 0.76 (0.19)$^{***}$ & 2.90 (2.94)$^{***}$ & 11.53 (14.92)$^{***}$ & 0.79 (0.20)$^{***}$ & 2.27 (1.89)$^{***}$ & 10.25 (11.76)$^{***}$ & 0.85 (0.12)$^{***}$ & 1.58 (1.37)$^{***}$ & 6.63 (7.19)$^{***}$ \\
\midrule
BCP \citep{bai2023bidirectional}   & \underline{0.70 (0.18)}$^{***}$ & 3.02 (1.80)$^{***}$ & \underline{10.34 (7.15)}$^{***}$ & 0.83 (0.13)$^{***}$ & 2.61 (2.35)$^{***}$ & 8.60 (8.10)$^{***}$ & 0.90 (0.08)$^{***}$ & 1.55 (2.04)$^{***}$ & 4.90 (6.43)$^{***}$ \\
CM  \citep{zhao2024crossmatch}     & 0.68 (0.73)$^{***}$ & \underline{2.25 (1.58)}$^{***}$ & 12.75 (16.53)$^{***}$ & \underline{0.86 (0.13)}$^{***}$ & \underline{1.47 (1.11)}$^{***}$ & \textbf{6.36 (7.26)} & \underline{0.91 (0.07)}$^{***}$ & \underline{1.08 (0.61)} & 4.30 (3.61) \\
CML  \citep{wu2024cross}           & 0.04 (0.08)$^{***}$ & 2.82 (4.27)$^{***}$ & 53.29 (21.70)$^{***}$ & 0.17 (0.18)$^{***}$ & 2.35 (1.26)$^{***}$ & 41.16 (20.01)$^{***}$ & \underline{0.91 (0.08)}$^{***}$ & 1.17 (0.87)$^{***}$ & \underline{4.24 (5.33)} \\
DyCON  \citep{assefa2025dycon}     & 0.42 (0.30)$^{***}$ & 3.48 (2.14)$^{***}$ & 21.19 (14.06)$^{***}$ & 0.38 (0.24)$^{***}$ & 3.53 (2.04)$^{***}$ & 24.15 (11.91)$^{***}$ & \underline{0.91 (0.08)}$^{**}$ & 1.15 (0.91)$^{***}$ & \textbf{4.12 (4.12)} \\
MagicNet \citep{chen2023magicnet}  & 0.52 (0.26)$^{***}$ & 18.63 (15.14)$^{***}$ & 48.75 (24.31)$^{***}$ & 0.57 (0.28)$^{***}$ & 15.84 (14.28)$^{***}$ & 44.30 (27.56)$^{***}$ & 0.87 (0.11)$^{***}$ & 1.83 (1.99)$^{***}$ & 6.48 (7.75)$^{***}$ \\
\midrule
RegAL (Ours)  & \textbf{0.86 (0.15)} & \textbf{1.69 (1.83)} & \textbf{7.33 (9.33)} & \textbf{0.87 (0.14)} & \textbf{1.30 (1.25)} & \underline{6.71 (8.82)} & \textbf{0.92 (0.08)} & \textbf{1.06 (0.86)} & \textbf{4.12 (5.36)} \\
\bottomrule
\end{tabular}
}
\end{table}

\begin{table}[!t]
\centering
\caption{Comparison of SSL Strategies on the ProstateX Dataset. We report Median (IQR) for Dice ($\uparrow$), ASD ($\downarrow$, mm), and HD95 ($\downarrow$, mm) across annotation budgets. \textbf{Bold} indicates the best value; \underline{underline} indicates the second-best among SSL methods. Superscripts denote statistical significance vs.\ RegAL (two-sided Wilcoxon signed-rank test, Holm--Bonferroni corrected): $^{*}p{<}0.05$, $^{**}p{<}0.01$, $^{***}p{<}0.001$.}
\label{tab:prostatex_summary_final}
\setlength{\tabcolsep}{2pt}
\resizebox{\textwidth}{!}{%
\begin{tabular}{l ccc | ccc | ccc}
\toprule
& \multicolumn{3}{c|}{\textbf{10 Labeled}} & \multicolumn{3}{c|}{\textbf{20 Labeled}} & \multicolumn{3}{c}{\textbf{100 Labeled}} \\
\textbf{Method} & \textbf{Dice} $\uparrow$ & \textbf{ASD} $\downarrow$ & \textbf{HD95} $\downarrow$ & \textbf{Dice} $\uparrow$ & \textbf{ASD} $\downarrow$ & \textbf{HD95} $\downarrow$ & \textbf{Dice} $\uparrow$ & \textbf{ASD} $\downarrow$ & \textbf{HD95} $\downarrow$ \\
\midrule
U-Net (Baseline) & 0.34 (0.44)$^{***}$ & 2.10 (4.09)$^{***}$ & 10.30 (14.27)$^{***}$ & 0.41 (0.39)$^{***}$ & 2.05 (4.66)$^{***}$ & 10.39 (19.67)$^{***}$ & 0.49 (0.39)$^{***}$ & 1.23 (1.08)$^{***}$ & 5.79 (8.26)$^{***}$ \\
\midrule
BCP \citep{bai2023bidirectional}   & \underline{0.40 (0.31)}$^{**}$  & 2.94 (3.82)$^{***}$ & 11.25 (14.09)$^{***}$ & \underline{0.44 (0.38)}$^{*}$  & 1.88 (2.61)$^{***}$ & 7.07 (10.44)          & 0.49 (0.35)$^{***}$ & 1.42 (1.12)$^{***}$ & 6.16 (8.08)$^{***}$ \\
CM \citep{zhao2024crossmatch}      & 0.19 (0.59)$^{***}$             & \underline{2.19 (3.60)}$^{***}$ & \underline{8.22 (7.94)}   & 0.34 (0.65)$^{***}$ & \underline{1.77 (1.97)}$^{*}$  & \underline{7.00 (8.22)} & 0.32 (0.71)$^{***}$ & 1.15 (1.20)$^{*}$   & \underline{5.00 (4.07)}$^{*}$ \\
CML \citep{wu2024cross}            & 0.10 (0.37)$^{***}$             & 2.74 (7.77)$^{***}$ & 20.58 (27.44)$^{***}$  & 0.11 (0.38)$^{***}$ & 2.09 (4.00)$^{***}$ & 21.93 (21.12)$^{***}$  & \underline{0.51 (0.36)}$^{**}$ & \textbf{1.02 (0.66)} & 5.48 (8.04)$^{***}$ \\
DyCON \citep{assefa2025dycon}      & 0.00 (0.15)$^{***}$             & 9.01 (10.03)$^{***}$ & 43.84 (22.95)$^{***}$ & 0.00 (0.16)$^{***}$ & 3.14 (4.11)$^{***}$ & 20.12 (19.93)$^{***}$  & 0.45 (0.36)$^{***}$ & 1.30 (1.03)$^{***}$ & 6.32 (8.54)$^{***}$ \\
MagicNet \citep{chen2023magicnet}  & \underline{0.40 (0.31)}$^{**}$  & 2.93 (2.80)$^{***}$ & 11.04 (10.76)$^{***}$  & 0.43 (0.36)$^{***}$ & 2.11 (2.01)$^{***}$ & 8.60 (9.45)$^{***}$    & 0.48 (0.34)$^{***}$ & 1.54 (1.34)$^{***}$ & 6.00 (6.54)$^{***}$ \\
\midrule
RegAL (Ours)  & \textbf{0.42 (0.40)} & \textbf{1.38 (1.18)} & \textbf{7.28 (9.91)} & \textbf{0.46 (0.39)} & \textbf{1.29 (1.22)} & \textbf{6.54 (9.23)} & \textbf{0.53 (0.36)} & \underline{1.07 (0.84)} & \textbf{4.26 (6.46)} \\
\bottomrule
\end{tabular}
}
\end{table}

\begin{table}[!t]
\centering
\caption{Comparison of SSL strategies on the dHCP neonatal brain dataset (Protocol 3). Median (IQR) Dice ($\uparrow$), ASD ($\downarrow$, mm), and HD95 ($\downarrow$, mm). \textbf{Bold}: best value; \underline{underline}: second-best among SSL methods. Superscripts denote significance vs.\ RegAL (two-sided Wilcoxon signed-rank test, Holm--Bonferroni corrected): $^{***}p{<}0.001$.}
\label{tab:dhcp_executive_summary_no_oracle}
\setlength{\tabcolsep}{3pt}
\resizebox{\textwidth}{!}{%
\begin{tabular}{l ccc | ccc}
\toprule
& \multicolumn{3}{c|}{\textbf{5 Labeled Samples}} & \multicolumn{3}{c}{\textbf{10 Labeled Samples}} \\
\textbf{Method} & \textbf{Dice} $\uparrow$ & \textbf{ASD} $\downarrow$ & \textbf{HD95} $\downarrow$ & \textbf{Dice} $\uparrow$ & \textbf{ASD} $\downarrow$ & \textbf{HD95} $\downarrow$ \\
\midrule
U-Net (Baseline)                       & 0.84 (0.19)$^{***}$             & 0.97 (1.55)$^{***}$ & 2.92 (4.67)$^{***}$             & 0.85 (0.14)$^{***}$             & 0.81 (1.22)$^{***}$ & 2.50 (3.58)$^{***}$ \\
\midrule
BCP      \citep{bai2023bidirectional}  & \underline{0.87 (0.11)}$^{***}$ & 0.60 (0.55)$^{***}$ & 1.58 (1.55)$^{***}$             & \underline{0.88 (0.10)}$^{***}$ & 0.56 (0.57)$^{***}$ & 1.50 (1.58)$^{***}$ \\
CM       \citep{zhao2024crossmatch}    & \underline{0.87 (0.12)}$^{***}$ & 0.64 (1.24)$^{***}$ & \underline{1.50 (4.22)}$^{***}$ & \textbf{0.89 (0.08)}$^{***}$    & 0.44 (0.34)$^{***}$ & \textbf{1.00 (0.79)} \\
CML      \citep{wu2024cross}           & \underline{0.87 (0.11)}$^{***}$ & \underline{0.46 (0.25)}$^{***}$ & \underline{1.50 (1.31)}$^{***}$ & \underline{0.88 (0.10)}$^{***}$ & \underline{0.43 (0.21)}$^{***}$ & \underline{1.22 (1.00)}$^{***}$ \\
DyCON    \citep{assefa2025dycon}       & 0.73 (0.25)$^{***}$             & 1.66 (2.39)$^{***}$ & 4.74 (6.87)$^{***}$             & 0.78 (0.18)$^{***}$             & 1.57 (1.86)$^{***}$ & 4.72 (5.95)$^{***}$ \\
MagicNet \citep{chen2023magicnet}      & 0.45 (0.56)$^{***}$             & 4.41 (6.96)$^{***}$ & 13.93 (18.32)$^{***}$           & 0.67 (0.46)$^{***}$             & 1.91 (5.17)$^{***}$ & 8.73 (15.52)$^{***}$ \\
\midrule
RegAL (Ours) & \textbf{0.88 (0.10)} & \textbf{0.43 (0.28)} & \textbf{1.12 (0.63)} & \textbf{0.89 (0.10)} & \textbf{0.39 (0.21)} & \textbf{1.00 (0.71)} \\
\bottomrule
\end{tabular}
}
\end{table}

In the ultra-low data regime (10 Labeled) of the BraTS and ProstateX experiments, our registration-guided Mean Teacher framework demonstrates clear advantages in stability and robustness. several consistency- and pseudo-label–based baselines exhibit substantial performance degradation or model collapse (near-zero Dice) when supervision is extremely limited. %In contrast, our method maintains anatomically coherent predictions and stable boundary delineation, reflecting the benefit of using diffeomorphic registration to generate geometrically valid supervision.
Particularly, we find the registration-guided SSL is also effective on BraTS, where inter-subject registration is usually considered unreliable due to different tumor locations and appearances. We believe three mechanisms in our method alleviate the impact of imperfect registration: (1) atlases selected through the Pareto criterion favour labeled samples with low uncertainty and high feature similarity to the target, significantly reducing the difficulty of registration; (2) rather than propagating warped labels onto the potentially mismatched target image, our augmentation jointly warps the atlas image and its corresponding label using a diffeomorphic deformation field, producing anatomically consistent training pairs regardless of alignment accuracy; and (3) the Mean-Teacher EMA buffers any residual noise in the augmented pairs. These mechanisms make our SSL more robust than the baselines in the ultra-low data regime.

%We note that the supervised U-Net baseline, which receives identical labeled data without any unlabeled data utilization, serves as a lower bound isolating the contribution of the SSL component as a whole. The consistent gap between RegAL and the U-Net baseline across all annotation budgets, most pronounced in the ultra-low regime where other SSL methods collapse, confirms that registration-guided augmentation provides stable and reliable supervisory signal precisely where standard consistency-based approaches fail.

As the annotation budget increases to a moderate level, performance gaps narrow across methods. On BraTS and ProstateX, when the labeled set becomes sufficiently large (100 Labeled), the performance of different SSL strategies converges and becomes broadly comparable. This trend is expected: as the amount of verified supervision increases, the relative contribution of SSL diminishes. Importantly, our method remains competitive in this regime, indicating that the proposed augmentation strategy does not impede learning when supervision is abundant.

Similar to the previous two testing protocols, we only test the low-budget scenarios for dHCP in protocol 3, where RegAL consistently achieves the best Dice and ASD across all tissue classes at both 5 and 10 labeled samples (Table~\ref{tab:dhcp_executive_summary_no_oracle}), with particularly pronounced gains on structurally complex and topologically vulnerable regions such as the Ventricles and Hippocampus (Supplementary Tables A8 \& A9). Qualitative segmentation maps on the dHCP dataset (Supplementary Fig. A4) further illustrate this pattern: while RegAL preserves topological coherence across all eight tissue classes at 5 labeled samples, competing methods exhibit progressive degradation on subcortical structures, with DyCON and MagicNet producing severely fragmented predictions on the Ventricles and Hippocampus, precisely the structures most vulnerable to cold-start collapse.

% === end hidden EMA Teacher ablation ===

\iffalse % === Discussion paragraph for the hidden EMA Teacher ablation; restore alongside Table~\ref{tab:student_teacher_ablation} if reinstated ===
The EMA Teacher provides temporal ensembling that stabilizes predictions and tightens anatomical boundaries. Consistent with the ablation in Table~\ref{tab:student_teacher_ablation}, its benefit is concentrated on the boundary metrics (ASD, HD95) and is largest when supervision is scarcest, while volumetric overlap (Dice) is essentially unchanged; once the labeled budget grows the Student and Teacher converge. This is expected: the moving-average Teacher acts as a spatial and temporal regularizer that smooths the prediction oscillations to which the raw Student is most susceptible under low-contrast, ambiguous boundaries (such as neonatal subcortical structures and low-contrast prostate zones) without altering the overall region estimate.
\fi % === end hidden EMA Discussion paragraph ===

\subsection{Hyperparameter robustness.} Lastly, we show that RegAL is robust to the choice of augmentation and topology weights. Specifically, we examine the Median Dice on the ProstateX dataset in the 100-labeled setting of Protocol 3. The performance varies by at most $\pm2\%$ across a 20$\times$ range of $\lambda_{\mathrm{conn}}$ and a 40$\times$ range of $\lambda_{\mathrm{TV}}$ (Table~\ref{tab:lambda_sweep}). This insensitivity is expected, since Pareto dominance and crowding-distance ranking are invariant to positive rescaling of any single objective; $\lambda_{\mathrm{conn}}$ therefore affects selection only through numerical conditioning of the topology axis relative to $\mathcal{H}$ and $\delta$. Notably, the optimal hyperparameter values selected in this particular setting are transferred directly across all datasets and annotation budgets without further tuning, confirming that the framework requires no dataset-specific calibration.

\begin{table}[!h]
\centering
\caption{Median Dice in the ProstateX Protocol~3, 100-labeled setting under a sweep of $\lambda_{\mathrm{conn}}$ and $\lambda_{\mathrm{TV}}$. Bold marks the default value used throughout.}
\label{tab:lambda_sweep}
\resizebox{0.8\textwidth}{!}{%
\begin{tabular}{lccccc|lccccc}
\toprule
\multicolumn{6}{c|}{\textbf{(a) $\lambda_{\mathrm{conn}}$}} & \multicolumn{5}{c}{\textbf{(b) $\lambda_{\mathrm{TV}}$}} \\
\cmidrule(r){1-6}\cmidrule(l){7-11}
 & 0.5 & 1.0 & \textbf{2.0} & 5.0 & 10.0 &  & 0.005 & 0.015 & \textbf{0.03} & 0.1 & 0.2 \\
\midrule
\textbf{Dice} & 0.53 & 0.54 & \textbf{0.53} & 0.55 & 0.55 & \textbf{Dice} & 0.55 & 0.54 & \textbf{0.53} & 0.53 & 0.55 \\
\bottomrule
\end{tabular}%
}
\end{table}

\subsection*{Limitations}
While RegAL demonstrates strong data efficiency across three publicly available benchmarks, several limitations should be acknowledged. First, the topological consistency metric $\mathcal{T}$ relies on the set $\mathcal{C}_{conn}$ of anatomical classes expected to form a single connected component. Although we construct this set automatically, the underlying assumption of topological simplicity does not hold for all anatomical structures (e.g., cortical grey matter folds, vascular trees). Future work will explore replacing this heuristic with learnable topological priors via persistent homology \citep{stucki2023topologically}. Second, the registration-guided augmentation assumes diffeomorphic deformability between subjects, which may be violated in the presence of severe anatomical missingness such as large post-operative resections; extending the framework to such cases is left for future work. Third, all three datasets are single-institution, publicly released cohorts, with no external or prospective validation performed. The generalization of RegAL to data acquired under significantly different scanner vendors, field strengths, or imaging protocols remains to be confirmed. %Fourth, owing to the computational cost of full 20{,}000-iteration 3D training, the higher-budget per-method comparisons (Protocols~1 and~3) report a single fixed-seed run rather than a complete cross-seed distribution; we mitigate this by stress-testing the most variance-prone ultra-low-budget regime across five independent random pools, where RegAL's advantage is preserved, but a full characterisation of cross-seed variance at every annotation budget remains future work. 
Finally, the absolute Dice scores on ProstateX, while consistently superior to all baselines, remain modest in the ultra-low regime; for clinical deployment, such performance would require further annotation investment beyond the budgets studied here. We note, however, that zonal prostate segmentation is intrinsically ambiguous: the achieved Dice is comparable to the inter-rater agreement reported for this dataset \citep{holmlund2024prostatezones}, indicating that a substantial fraction of the residual error reflects genuine annotation uncertainty at low-contrast zonal boundaries rather than model failure.

\section{Conclusion}
We introduced RegAL, a unified active semi-supervised framework that tackles the cold-start problem in medical image segmentation by governing both annotation and unlabeled-data utilization with a single topology-aware Pareto criterion over voxel-wise uncertainty, feature diversity, and topological consistency. The same criterion surfaces informative edge cases for labeling and stable atlases for registration-guided augmentation, so acquisition and unlabeled-data use are aligned within one coherent mechanism rather than stacked as independently designed modules. Across three diverse benchmarks (BraTS 2021, dHCP, and ProstateX), RegAL remained stable and data-efficient under extreme annotation scarcity and consistently outperformed state-of-the-art active, semi-supervised, and active semi-supervised baselines on both overlap and boundary-distance metrics. These results indicate that anatomical structure and geometric priors are decisive ingredients for learning from very few labels, and that coupling acquisition with unlabeled-data utilization through one principled criterion is more effective than stacking heuristic modules.

%Looking ahead, a natural extension is to integrate RegAL with foundation models pre-trained on large-scale 3D medical imaging, such as those built on the Segment Anything paradigm. RegAL is backbone-agnostic by design: its Pareto selection operates on the features and predictions of any segmentation network, so pairing it with a strongly pre-trained backbone would let the uncertainty and diversity signals reflect residual rather than random-initialization uncertainty, potentially accelerating cold-start convergence further. We leave this integration, together with the learnable topological priors and resection-aware deformation models discussed above, to future work.

\section*{Data Availability}
All datasets used in this study are publicly available. 
The BraTS 2021 dataset is available through the RSNA-ASNR-MICCAI 
Brain Tumor Segmentation Challenge at 
\url{https://www.synapse.org/\#!Synapse:syn27046444}. 
The Developing Human Connectome Project (dHCP) dataset is available 
at \url{https://www.developingconnectome.org}. 
The ProstateX zonal segmentation dataset is available through 
The Cancer Imaging Archive at 
\url{https://www.cancerimagingarchive.net/collection/prostatex}. 
\section*{Code Availability}
The source code for RegAL is available at \url{https://github.com/BahramJafrasteh/Regal}.

\section*{Acknowledgments}
This work was supported in part by NAIRR Pilot Grant
NAIRR250120 (to Q.Z.). 
\section*{Declaration of Competing Interests}
The authors declare that they have no known competing financial
interests or personal relationships that could have appeared to
influence the work reported in this paper.

\section*{CRediT authorship contribution statement}
\textbf{Bahram Jafrasteh}: Conceptualization, Methodology, Software, Validation, Formal analysis, Investigation, Data curation, Original draft writing, Review and editing, Visualization.
\textbf{Cheng Wan}: Software, Formal analysis, Review and editing.
\textbf{Heejong Kim}: Review and editing.
\textbf{Johannes C. Paetzold}: Review and editing.
\textbf{Qingyu Zhao}: Conceptualization, Methodology, Supervision, Project administration, Original draft writing, Review and editing.

\clearpage

\appendix
\setcounter{table}{0}
\renewcommand{\thetable}{A\arabic{table}}
\setcounter{figure}{0}
\renewcommand{\thefigure}{A\arabic{figure}}
\section*{Supplementary Material}

This supplement provides detailed per-class and per-zone breakdowns of results summarised in the main paper, together with the complete hyperparameter specification and full qualitative comparisons.
The supplement is organised by protocol, in the order these results are first discussed in the main text: Table~\ref{tab:hyperparams} gives the complete hyperparameter specification (the hyperparameter sensitivity sweep itself is reported in the main text, Table~8); Tables~\ref{tab:approach3_prostateX} and \ref{tab:dhcp_assl_perclass} detail the Protocol~1 comparison on ProstateX and dHCP, respectively, and Fig.~\ref{fig:qualitative_sagittal} provides a sagittal-plane counterpart to the main-text qualitative comparison; Tables~\ref{tab:al_comparison_dice_iqr} and \ref{tab:dhcp_al_perclass} detail the Protocol~2 comparison on ProstateX and dHCP; and Tables~\ref{tab:prostatex_dice}--\ref{tab:dhcp_asd_only} together with Figs.~\ref{fig:prostate_qualitative}--\ref{fig:dhcp_qualitative} detail the Protocol~3 comparison and qualitative results across all three datasets.

\clearpage

\subsection*{A.\; Complete Hyperparameter Specification}

Table~\ref{tab:hyperparams} consolidates all model, training, and hyperparameter configurations for RegAL, shared across all three datasets and annotation budgets unless otherwise noted.

\begin{table}[!h]
\centering
\caption{Complete hyperparameter specification for RegAL.}
\label{tab:hyperparams}
\setlength{\tabcolsep}{6pt}
\begin{tabular}{llc}
\toprule
\textbf{Group} & \textbf{Hyperparameter} & \textbf{Value} \\
\midrule
\multirow{5}{*}{Segmentation} & Backbone & 3D U-Net \\
 & Encoder channels & 16, 32, 64, 128, 256 \\
 & Normalization / activation & Group Norm / Leaky ReLU \\
 & Optimizer & Adam \\
 & Learning rate & $10^{-4}$ \\
\midrule
\multirow{5}{*}{Training} & Total iterations & 20{,}000 \\
 & EMA decay ($\alpha$) & 0.999 \\
 & Warm-start per AL cycle & 200 iter \\
 & Consistency ramp-up & 0 $\to$ 1 over first 2{,}000 iter \\
 & AL/SSL multi-objective solver & NSGA-II \citep{deb2002fast} \\
\midrule
\multirow{4}{*}{Loss weights} & $\mathcal{L}_{focal},\,\mathcal{L}_{dice},\,\mathcal{L}_{cons}$ & 1.0 (unit) \\
 & $\lambda_{TV}$ & 0.03 \\
 & $\lambda_{conn}$ & 2.0 \\
 & Focal loss $\gamma$ & 2.0 \\
\midrule
\multirow{3}{*}{Registration} & Architecture & VoxelMorph (SVF, 7 steps) \\
 & $\lambda_{smooth}$, $\lambda_{jac}$ & 0.1, 0.1 \\
 & Learning rate & $10^{-4}$ \\
\midrule
\multirow{2}{*}{Volume size} & BraTS 2021, dHCP & $96\times96\times96$ \\
 & ProstateX & $128\times128\times32$ \\
\bottomrule
\end{tabular}
\end{table}

\subsection*{B.\; Protocol~1: Detailed Per-Region and Per-Class Results}

Table~\ref{tab:approach3_prostateX} extends the main-paper Protocol~1 results with per-zone detail on ProstateX across three annotation budgets (30, 50, 100).
The prostate zones vary substantially in size and boundary contrast --- the Transition Zone (TZ) is the largest and easiest, while the Urethra and Anterior Stroma (AS) are small, irregular structures that most baselines fail to segment at low label budgets. RegAL is the only method that produces consistent non-zero predictions for these minor zones even at the smallest budget tested.

\begin{table}[ht]
\centering
\caption{Comparing \textbf{regional} segmentation performance of ASSL frameworks on ProstateX across low (30), medium (50), and high (100) annotation budgets. RegAL consistently outperforms baselines, particularly on challenging minor structures. Values are reported as Median (IQR).}
\label{tab:approach3_prostateX}
\resizebox{\textwidth}{!}{
\begin{tabular}{l|c|ccccc|c}
\hline
\textbf{Budget} & \textbf{Method} & \textbf{PZ} & \textbf{CZ} & \textbf{TZ} & \textbf{AS} & \textbf{Urethra} & \textbf{Average Dice ($\uparrow$)} \\ \hline
\multirow{3}{*}{\textbf{30 Samples}}
 & AS3L \citep{wen2025active} & $0.47 (0.23)$ & $0.11 (0.20)$ & $0.77 (0.12)$ & $0.11 (0.27)$ & $0.00 (0.00)$ & $0.21 (0.55)$ \\
 & S4AL \citep{rangnekar2023semantic} & $0.48 (0.19)$ & $0.15 (0.26)$ & $0.77 (0.16)$ & $0.13 (0.24)$ & $0.00 (0.00)$ & $0.23 (0.54)$ \\
 & \textbf{RegAL (Ours)} & \textbf{0.69 (0.15)} & \textbf{0.44 (0.17)} & \textbf{0.83 (0.13)} & \textbf{0.38 (0.20)} & \textbf{0.39 (0.17)} & \textbf{0.49 (0.35)} \\ \hline
\multirow{3}{*}{\textbf{50 Samples}}
 & AS3L \citep{wen2025active} & $0.49 (0.21)$ & $0.15 (0.20)$ & $0.76 (0.12)$ & $0.16 (0.30)$ & $0.00 (0.05)$ & $0.26 (0.55)$ \\
 & S4AL \citep{rangnekar2023semantic} & $0.49 (0.21)$ & $0.12 (0.26)$ & $0.77 (0.16)$ & $0.09 (0.22)$ & $0.00 (0.00)$ & $0.20 (0.55)$ \\
 & \textbf{RegAL (Ours)} & \textbf{0.71 (0.14)} & \textbf{0.44 (0.14)} & \textbf{0.82 (0.11)} & \textbf{0.36 (0.22)} & \textbf{0.40 (0.14)} & \textbf{0.52 (0.36)} \\ \hline
\multirow{3}{*}{\textbf{100 Samples}}
 & AS3L \citep{wen2025active} & $0.54 (0.17)$ & $0.29 (0.24)$ & $0.77 (0.12)$ & $0.22 (0.30)$ & $0.07 (0.21)$ & $0.35 (0.47)$ \\
 & S4AL \citep{rangnekar2023semantic} & $0.52 (0.23)$ & $0.23 (0.27)$ & $0.77 (0.13)$ & $0.15 (0.28)$ & $0.00 (0.12)$ & $0.33 (0.53)$ \\
 & \textbf{RegAL (Ours)} & \textbf{0.73 (0.14)} & \textbf{0.50 (0.17)} & \textbf{0.84 (0.10)} & \textbf{0.39 (0.22)} & \textbf{0.40 (0.13)} & \textbf{0.54 (0.36)} \\ \hline
\end{tabular}
}
\end{table}

On dHCP, Table~\ref{tab:dhcp_assl_perclass} provides the full per-class Dice breakdown for the ASSL comparison across all 5 annotation budgets (1, 3, 5, 7, 9 labeled scans). The aggregate results (Dice, ASD, HD95) are summarised in the main text Table~3. The per-class view confirms that RegAL's largest gains occur on the topologically challenging Ventricle and Hippocampus structures, where ASSL baselines produce near-zero Dice in the ultra-low-label regime.

\begin{table}[!h]
\centering
\caption{Per-class Dice ($\uparrow$) of ASSL frameworks across annotation budgets on dHCP, averaged over 5 random seeds. \textbf{Abbrev.:} Vent (Ventricle), Cer (Cerebellum), BS (Brainstem), Hippo (Hippocampus). Values reported as Median (IQR). \textbf{Bold}: best per column.}
\label{tab:dhcp_assl_perclass}
\resizebox{\textwidth}{!}{%
\begin{tabular}{l|c|cccccccc|c}
\toprule
\textbf{Budget} & \textbf{Method} & \textbf{CSF} & \textbf{CGM} & \textbf{WM} & \textbf{Vent} & \textbf{Cer} & \textbf{DGM} & \textbf{BS} & \textbf{Hippo} & \textbf{Avg Dice ($\uparrow$)} \\
\midrule
\multirow{3}{*}{\textbf{1 Labeled}}
 & AS3L \citep{wen2025active}         & 0.70 (0.10) & 0.80 (0.05) & 0.82 (0.05) & 0.00 (0.18) & 0.79 (0.10) & 0.70 (0.06) & 0.56 (0.18) & 0.00 (0.01) & 0.69 (0.45) \\
 & S4AL \citep{rangnekar2023semantic} & 0.70 (0.07) & 0.80 (0.04) & 0.82 (0.05) & 0.13 (0.19) & 0.78 (0.13) & 0.69 (0.06) & 0.51 (0.27) & 0.00 (0.03) & 0.68 (0.52) \\
 & RegAL (Ours)                       & \textbf{0.70 (0.09)} & \textbf{0.77 (0.05)} & \textbf{0.78 (0.06)} & \textbf{0.32 (0.16)} & \textbf{0.83 (0.14)} & \textbf{0.76 (0.07)} & \textbf{0.64 (0.20)} & \textbf{0.27 (0.22)} & \textbf{0.72 (0.31)} \\
\midrule
\multirow{3}{*}{\textbf{3 Labeled}}
 & AS3L \citep{wen2025active}         & 0.74 (0.06) & 0.81 (0.05) & 0.85 (0.06) & 0.18 (0.31) & 0.85 (0.06) & 0.74 (0.06) & 0.62 (0.22) & 0.00 (0.22) & 0.74 (0.37) \\
 & S4AL \citep{rangnekar2023semantic} & 0.70 (0.07) & 0.78 (0.05) & 0.87 (0.04) & 0.06 (0.22) & 0.82 (0.08) & 0.82 (0.05) & 0.57 (0.24) & 0.00 (0.00) & 0.74 (0.44) \\
 & RegAL (Ours)                       & \textbf{0.81 (0.04)} & \textbf{0.86 (0.03)} & \textbf{0.90 (0.04)} & \textbf{0.74 (0.08)} & \textbf{0.92 (0.05)} & \textbf{0.89 (0.02)} & \textbf{0.91 (0.03)} & \textbf{0.72 (0.08)} & \textbf{0.86 (0.12)} \\
\midrule
\multirow{3}{*}{\textbf{5 Labeled}}
 & AS3L \citep{wen2025active}         & 0.75 (0.05) & 0.81 (0.05) & 0.84 (0.06) & 0.31 (0.19) & 0.86 (0.06) & 0.74 (0.06) & 0.61 (0.23) & 0.13 (0.28) & 0.74 (0.37) \\
 & S4AL \citep{rangnekar2023semantic} & 0.76 (0.06) & 0.83 (0.04) & 0.89 (0.04) & 0.45 (0.38) & 0.86 (0.06) & 0.84 (0.03) & 0.73 (0.11) & 0.00 (0.03) & 0.80 (0.20) \\
 & RegAL (Ours)                       & \textbf{0.82 (0.04)} & \textbf{0.86 (0.04)} & \textbf{0.90 (0.04)} & \textbf{0.76 (0.08)} & \textbf{0.93 (0.06)} & \textbf{0.91 (0.02)} & \textbf{0.92 (0.02)} & \textbf{0.77 (0.06)} & \textbf{0.87 (0.11)} \\
\midrule
\multirow{3}{*}{\textbf{7 Labeled}}
 & AS3L \citep{wen2025active}         & 0.76 (0.05) & 0.81 (0.05) & 0.85 (0.06) & 0.33 (0.16) & 0.87 (0.05) & 0.75 (0.06) & 0.61 (0.24) & 0.22 (0.28) & 0.75 (0.36) \\
 & S4AL \citep{rangnekar2023semantic} & 0.79 (0.05) & 0.84 (0.04) & 0.89 (0.03) & 0.56 (0.26) & 0.88 (0.05) & 0.85 (0.03) & 0.78 (0.09) & 0.03 (0.16) & 0.82 (0.16) \\
 & RegAL (Ours)                       & \textbf{0.84 (0.03)} & \textbf{0.87 (0.03)} & \textbf{0.91 (0.03)} & \textbf{0.81 (0.06)} & \textbf{0.94 (0.03)} & \textbf{0.92 (0.02)} & \textbf{0.92 (0.02)} & \textbf{0.79 (0.05)} & \textbf{0.88 (0.10)} \\
\midrule
\multirow{3}{*}{\textbf{9 Labeled}}
 & AS3L \citep{wen2025active}         & 0.76 (0.04) & 0.81 (0.05) & 0.85 (0.06) & 0.32 (0.16) & 0.88 (0.05) & 0.75 (0.06) & 0.60 (0.25) & 0.25 (0.26) & 0.76 (0.36) \\
 & S4AL \citep{rangnekar2023semantic} & 0.80 (0.05) & 0.85 (0.04) & 0.90 (0.03) & 0.61 (0.19) & 0.88 (0.05) & 0.86 (0.03) & 0.81 (0.08) & 0.16 (0.29) & 0.83 (0.15) \\
 & RegAL (Ours)                       & \textbf{0.84 (0.03)} & \textbf{0.87 (0.03)} & \textbf{0.91 (0.03)} & \textbf{0.81 (0.06)} & \textbf{0.95 (0.03)} & \textbf{0.92 (0.02)} & \textbf{0.93 (0.02)} & \textbf{0.79 (0.05)} & \textbf{0.88 (0.09)} \\
\bottomrule
\end{tabular}
}
\end{table}

\begin{figure*}[t]
    \centering
    \includegraphics[width=\textwidth]{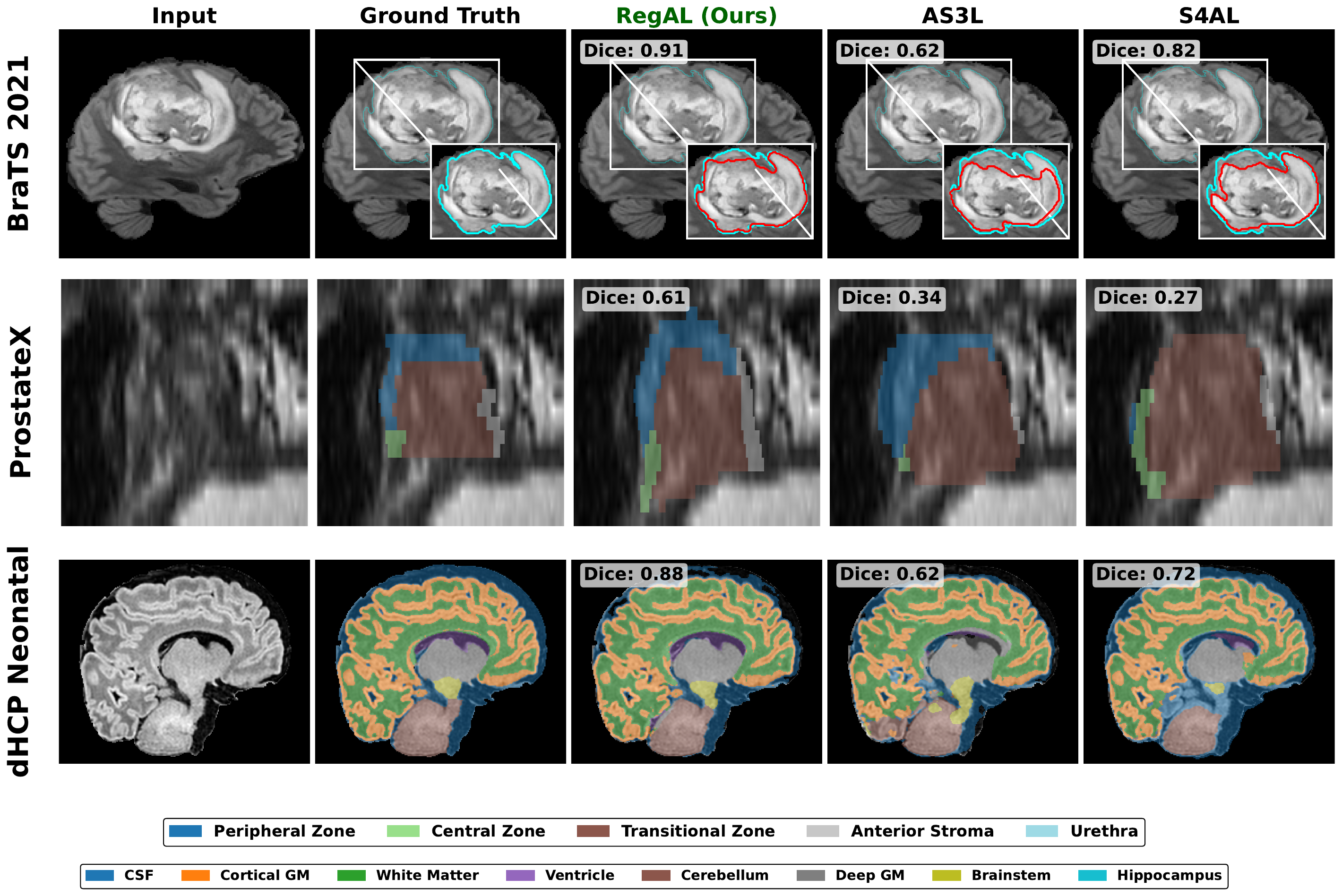}
    \caption{\textbf{Sagittal-plane qualitative comparison, complementing main-text Fig.~3.} Same cases, methods, and annotation budgets as the main-text axial comparison (BraTS: 50 labeled samples; ProstateX: 20 labeled samples; dHCP: 9 labeled samples), shown here in the sagittal plane. RegAL (Ours) maintains its qualitative advantage in boundary adherence and topological coherence across this orthogonal view, consistent with the axial results.}
    \label{fig:qualitative_sagittal}
\end{figure*}

Fig.~\ref{fig:qualitative_sagittal} shows the same cases as the main-text qualitative comparison (Fig.~3) in the sagittal plane, confirming that RegAL's advantage in boundary adherence and topological coherence holds across orthogonal views rather than being an artefact of the single plane shown in the main text.

\subsection*{C.\; Protocol~2: Detailed Per-Region and Per-Class Results}

Table~\ref{tab:al_comparison_dice_iqr} reports the Protocol~2 AL-strategy comparison on ProstateX, broken down by prostate zone.
Single-metric heuristics (Uncertainty, Diversity, Topology) show inconsistent per-zone behaviour --- performing well on the dominant TZ while failing entirely on minor structures --- whereas Pareto($\mathcal{H}$,$\mathcal{T}$,$\delta$) yields non-zero Dice across all zones at every budget.

\begin{table}[H]
\centering
\caption{Comparison of \textbf{regional} segmentation accuracy across different AL strategies. Values are reported as \textbf{Median (IQR)}. \textbf{Bold} indicates the best performance in each category. The proposed multi-objective  strategy based on Pareto($\mathcal{H}$, $\mathcal{T}$, $\delta$) demonstrates competitive performance, particularly in the Peripheral and Transitional Zones.}
\label{tab:al_comparison_dice_iqr}
\resizebox{\textwidth}{!}{%
\begin{tabular}{lcccccc}
\toprule
\textbf{Method} & \textbf{PZ} & \textbf{CZ} & \textbf{TZ} & \textbf{AS} & \textbf{Urethra} & \textbf{Average} \\
\midrule
% --- 3 LABELED ---
\multicolumn{7}{c}{\textbf{\textit{3 Labeled Samples}}} \\
\midrule
Random & 0.35 (0.25) & 0.16 (0.15) & 0.54 (0.31) & \textbf{0.10 (0.16)} & \textbf{0.10 (0.13)} & 0.19 (0.28) \\
Uncertainty \citep{gal2016dropout} & 0.36 (0.28) & 0.02 (0.07) & 0.55 (0.35) & 0.06 (0.22) & 0.06 (0.22) & 0.17 (0.41) \\
Diversity \citep{sener2018active} ($\delta$) & 0.06 (0.16) & 0.00 (0.00) & 0.10 (0.41) & 0.00 (0.03) & 0.00 (0.00) & 0.00 (0.10) \\
Topology $\mathcal{T}$ & \textbf{0.45 (0.38)} & 0.04 (0.19) & \textbf{0.61 (0.30)} & 0.07 (0.19) & 0.04 (0.16) & 0.18 (0.43) \\
Pareto($\mathcal{H}$, $\mathcal{T}$, $\delta$) & 0.41 (0.32) & \textbf{0.21 (0.26)} & 0.52 (0.37) & \textbf{0.10 (0.20)} & 0.02 (0.09) & \textbf{0.21 (0.40)} \\
\midrule

% --- 5 LABELED ---
\multicolumn{7}{c}{\textbf{\textit{5 Labeled Samples}}} \\
\midrule
Random & 0.38 (0.25) & \textbf{0.23 (0.18)} & 0.60 (0.30) & 0.14 (0.15) & 0.12 (0.15) & 0.24 (0.30) \\
Uncertainty \citep{gal2016dropout} & 0.40 (0.33) & 0.11 (0.21) & 0.68 (0.25) & 0.01 (0.13) & 0.01 (0.11) & 0.19 (0.45) \\
Diversity \citep{sener2018active} ($\delta$) & 0.23 (0.34) & 0.07 (0.21) & 0.56 (0.36) & \textbf{0.17 (0.31)} & 0.06 (0.25) & 0.21 (0.37) \\
Topology $\mathcal{T}$ & 0.47 (0.32) & 0.14 (0.26) & \textbf{0.71 (0.17)} & 0.12 (0.31) & 0.08 (0.21) & 0.27 (0.50) \\
Pareto($\mathcal{H}$, $\mathcal{T}$, $\delta$) & \textbf{0.48 (0.36)} & \textbf{0.23 (0.30)} & 0.70 (0.22) & 0.16 (0.28) & \textbf{0.15 (0.26)} & \textbf{0.30 (0.44)} \\
\midrule

% --- 7 LABELED ---
\multicolumn{7}{c}{\textbf{\textit{7 Labeled Samples}}} \\
\midrule
Random & 0.47 (0.25) & 0.26 (0.18) & 0.68 (0.26) & 0.18 (0.18) & 0.11 (0.13) & 0.27 (0.37) \\
Uncertainty \citep{gal2016dropout} & 0.43 (0.29) & 0.25 (0.24) & 0.73 (0.18) & 0.14 (0.28) & 0.00 (0.08) & 0.27 (0.45) \\
Diversity \citep{sener2018active} ($\delta$) & \textbf{0.50 (0.32)} & \textbf{0.27 (0.32)} & 0.67 (0.26) & 0.12 (0.28) & 0.05 (0.20) & 0.28 (0.48) \\
Topology $\mathcal{T}$ & 0.45 (0.34) & 0.14 (0.27) & 0.72 (0.13) & 0.16 (0.28) & 0.06 (0.28) & 0.28 (0.49) \\
Pareto($\mathcal{H}$, $\mathcal{T}$, $\delta$) & \textbf{0.50 (0.39)} & 0.22 (0.27) & \textbf{0.75 (0.24)} & \textbf{0.19 (0.23)} & \textbf{0.12 (0.31)} & \textbf{0.31 (0.44)} \\
\bottomrule
\end{tabular}%
}
\end{table}

On dHCP, Table~\ref{tab:dhcp_al_perclass} provides the full per-class Dice breakdown for the AL-only comparison (Protocol~2) across all three annotation budgets (3, 5, 7 labeled volumes). The aggregate results (Dice, ASD, HD95) are summarised in the main text (Table~5). The per-class view shows that the Pareto objective's advantage at $N{=}3$ is broad-based across tissue classes, including the topologically challenging Ventricle and Hippocampus, and that the convergence among strategies at $N{=}5$ and $N{=}7$ is likewise consistent across classes rather than driven by any single structure.

\begin{table}[!h]
\centering
\caption{Per-class Dice ($\uparrow$) of AL strategies across annotation budgets on dHCP (Protocol~2), aggregated over all subjects. \textbf{Abbrev.:} Vent (Ventricle), Cer (Cerebellum), BS (Brainstem), Hippo (Hippocampus). Values reported as Median (IQR). \textbf{Bold}: best per column.}
\label{tab:dhcp_al_perclass}
\resizebox{\textwidth}{!}{%
\begin{tabular}{l|c|cccccccc|c}
\toprule
\textbf{Budget} & \textbf{Method} & \textbf{CSF} & \textbf{CGM} & \textbf{WM} & \textbf{Vent} & \textbf{Cer} & \textbf{DGM} & \textbf{BS} & \textbf{Hippo} & \textbf{Avg Dice ($\uparrow$)} \\
\midrule
\multirow{5}{*}{\textbf{3 Labeled}}
 & Random & 0.74 (0.05) & 0.78 (0.04) & 0.84 (0.05) & 0.60 (0.17) & 0.91 (0.04) & \textbf{0.89 (0.04)} & 0.86 (0.07) & 0.66 (0.14) & 0.79 (0.17) \\
 & Uncertainty \citep{gal2016dropout} & \textbf{0.75 (0.05)} & \textbf{0.80 (0.04)} & 0.84 (0.05) & \textbf{0.65 (0.11)} & 0.91 (0.04) & 0.88 (0.03) & 0.88 (0.04) & 0.66 (0.12) & 0.81 (0.15) \\
 & Diversity \citep{sener2018active} & 0.70 (0.07) & 0.73 (0.09) & 0.80 (0.06) & 0.57 (0.12) & 0.91 (0.03) & 0.87 (0.04) & 0.85 (0.07) & 0.65 (0.12) & 0.76 (0.20) \\
 & Topology ($\mathcal{T}$) & \textbf{0.75 (0.04)} & \textbf{0.80 (0.04)} & 0.84 (0.05) & 0.64 (0.11) & 0.91 (0.03) & 0.88 (0.03) & 0.87 (0.05) & 0.65 (0.15) & 0.81 (0.15) \\
 & Pareto (Ours) & \textbf{0.75 (0.04)} & \textbf{0.80 (0.04)} & \textbf{0.85 (0.04)} & \textbf{0.65 (0.11)} & \textbf{0.93 (0.02)} & \textbf{0.89 (0.03)} & \textbf{0.89 (0.04)} & \textbf{0.71 (0.09)} & \textbf{0.82 (0.16)} \\
\midrule
\multirow{5}{*}{\textbf{5 Labeled}}
 & Random & \textbf{0.78 (0.04)} & \textbf{0.82 (0.03)} & \textbf{0.87 (0.04)} & \textbf{0.71 (0.13)} & 0.93 (0.03) & \textbf{0.91 (0.03)} & \textbf{0.90 (0.05)} & \textbf{0.74 (0.11)} & \textbf{0.83 (0.13)} \\
 & Uncertainty \citep{gal2016dropout} & \textbf{0.78 (0.05)} & 0.81 (0.04) & 0.85 (0.05) & 0.67 (0.13) & 0.92 (0.03) & 0.90 (0.03) & 0.89 (0.04) & 0.70 (0.11) & 0.82 (0.14) \\
 & Diversity \citep{sener2018active} & \textbf{0.78 (0.05)} & 0.81 (0.06) & 0.86 (0.05) & 0.68 (0.11) & \textbf{0.94 (0.02)} & 0.90 (0.03) & \textbf{0.90 (0.04)} & \textbf{0.74 (0.07)} & 0.82 (0.14) \\
 & Topology ($\mathcal{T}$) & \textbf{0.78 (0.05)} & 0.80 (0.05) & 0.85 (0.05) & 0.66 (0.15) & 0.92 (0.03) & 0.90 (0.03) & 0.89 (0.04) & 0.71 (0.11) & 0.82 (0.15) \\
 & Pareto (Ours) & \textbf{0.78 (0.04)} & 0.81 (0.04) & 0.86 (0.05) & 0.69 (0.11) & 0.93 (0.02) & \textbf{0.91 (0.02)} & \textbf{0.90 (0.03)} & \textbf{0.74 (0.08)} & \textbf{0.83 (0.14)} \\
\midrule
\multirow{5}{*}{\textbf{7 Labeled}}
 & Random & \textbf{0.79 (0.03)} & \textbf{0.84 (0.03)} & \textbf{0.88 (0.04)} & \textbf{0.74 (0.10)} & \textbf{0.94 (0.02)} & \textbf{0.92 (0.03)} & \textbf{0.91 (0.04)} & 0.75 (0.09) & \textbf{0.85 (0.13)} \\
 & Uncertainty \citep{gal2016dropout} & \textbf{0.79 (0.04)} & 0.81 (0.04) & 0.86 (0.05) & 0.67 (0.11) & 0.92 (0.02) & 0.90 (0.03) & 0.89 (0.03) & 0.68 (0.13) & 0.83 (0.14) \\
 & Diversity \citep{sener2018active} & \textbf{0.79 (0.04)} & 0.83 (0.05) & 0.87 (0.04) & 0.73 (0.09) & 0.92 (0.03) & 0.91 (0.03) & \textbf{0.91 (0.03)} & \textbf{0.76 (0.06)} & 0.82 (0.14) \\
 & Topology ($\mathcal{T}$) & \textbf{0.79 (0.04)} & 0.81 (0.04) & 0.86 (0.05) & 0.65 (0.12) & 0.92 (0.03) & 0.89 (0.03) & 0.88 (0.04) & 0.69 (0.12) & \textbf{0.84 (0.14)} \\
 & Pareto (Ours) & \textbf{0.79 (0.04)} & \textbf{0.84 (0.03)} & \textbf{0.88 (0.05)} & 0.73 (0.09) & 0.93 (0.02) & \textbf{0.92 (0.02)} & \textbf{0.91 (0.03)} & \textbf{0.76 (0.06)} & \textbf{0.85 (0.13)} \\
\bottomrule
\end{tabular}
}
\end{table}

\subsection*{D.\; Protocol~3: Detailed Per-Region, Per-Class, and Qualitative Results}

On ProstateX, Tables~\ref{tab:prostatex_dice} and \ref{tab:prostatex_asd_complete} expand the Protocol~3 summary (main paper Table~5) with full per-zone Dice and ASD, respectively.
Notably, CrossMatch (CM) and CML produce N/A entries for the Central Zone and Urethra at several budgets --- corresponding to zero predicted foreground --- confirming the cold-start collapse reported in the main text.

\begin{table}[H]
\centering
\caption{Comparison Dice scores ($\uparrow$) of SSL strategies on ProstateX data using 10, 20 and 100 labeled data. We compare methods across different labeled data regimes. \textbf{Bold} indicates the best performance; \underline{underline} indicates the second-best.}
\label{tab:prostatex_dice}
\resizebox{\textwidth}{!}{%
\begin{tabular}{lcccccc}
\toprule
\textbf{Method} & \textbf{PZ} & \textbf{CZ} & \textbf{TZ} & \textbf{AS} & \textbf{Urethra} & \textbf{Average} \\
\midrule
% --- 10 LABELED ---
\multicolumn{7}{c}{\textbf{\textit{Experiment 1: 10 Labeled Samples}}} \\
\midrule
U-Net (Baseline) & \underline{0.55 (0.24)} & 0.27 (0.22) & \underline{0.77 (0.17)} & 0.18 (0.23) & 0.23 (0.23) & 0.34 (0.44) \\
BCP \citep{bai2023bidirectional}            & 0.54 (0.18) & 0.30 (0.14) & 0.74 (0.19) & \textbf{0.31 (0.24)} & \textbf{0.30 (0.24)} & 0.40 (0.31) \\
CM \citep{zhao2024crossmatch}             & 0.54 (0.17) & 0.00 (0.00) & 0.76 (0.14) & 0.22 (0.21) & 0.00 (0.00) & 0.19 (0.59) \\
CML \citep{wu2024cross}            & 0.31 (0.46) & 0.00 (0.07) & 0.57 (0.73) & 0.05 (0.29) & 0.10 (0.25) & 0.10 (0.37) \\
DyCON \citep{assefa2025dycon}          & 0.18 (0.09) & 0.00 (0.00) & 0.15 (0.05) & 0.00 (0.00) & 0.00 (0.00) & 0.00 (0.15) \\
MagicNet \citep{chen2023magicnet}       & 0.52 (0.17) & 0.35 (0.18) & 0.74 (0.13) & \textbf{0.31 (0.30)} & 0.27 (0.11) & 0.40 (0.31) \\
RegAL (Ours)   & \textbf{0.62 (0.22)} & \textbf{0.36 (0.21)} & \textbf{0.80 (0.14)} & \textbf{0.31 (0.30)} & \underline{0.29 (0.21)} & \textbf{0.42 (0.40)} \\
\midrule

% --- 20 LABELED ---
\multicolumn{7}{c}{\textbf{\textit{Experiment 2: 20 Labeled Samples}}} \\
\midrule
U-Net (Baseline) & 0.62 (0.24) & 0.35 (0.23) & 0.75 (0.22) & 0.23 (0.27) & 0.32 (0.28) & 0.41 (0.39) \\
BCP \citep{bai2023bidirectional}            & \underline{0.64 (0.22)} & \textbf{0.38 (0.17)} & 0.79 (0.15) & \underline{0.27 (0.35)} & 0.32 (0.22) & 0.44 (0.38) \\
CM \citep{zhao2024crossmatch}             & 0.61 (0.17) & \textbf{0.38 (0.17)} & \underline{0.81 (0.14)} & 0.00 (0.00) & 0.00 (0.00) & 0.34 (0.65) \\
CML \citep{wu2024cross}            & 0.38 (0.43) & 0.00 (0.14) & 0.26 (0.51) & 0.00 (0.00) & 0.31 (0.29) & 0.11 (0.38) \\
DyCON \citep{assefa2025dycon}          & 0.12 (0.13) & 0.00 (0.01) & 0.24 (0.13) & 0.00 (0.00) & 0.00 (0.00) & 0.00 (0.16) \\
MagicNet \citep{chen2023magicnet}       & 0.59 (0.15) & \textbf{0.38 (0.21)} & 0.77 (0.13) & 0.24 (0.25) & 0.31 (0.13) & 0.43 (0.36) \\
RegAL (Ours)   & \textbf{0.66 (0.19)} & \underline{0.37 (0.19)} & \textbf{0.82 (0.14)} & \textbf{0.36 (0.28)} & \textbf{0.33 (0.18)} & \textbf{0.46 (0.39)} \\
\midrule

% --- 100 LABELED ---
\multicolumn{7}{c}{\textbf{\textit{Experiment 3: 100 Labeled Samples}}} \\
\midrule
U-Net (Oracle) & 0.70 (0.19) & 0.43 (0.15) & 0.83 (0.09) & 0.35 (0.19) & 0.40 (0.19) & 0.49 (0.39) \\
BCP \citep{bai2023bidirectional}              & 0.67 (0.19) & \underline{0.45 (0.13)} & 0.81 (0.10) & \underline{0.38 (0.30)} & 0.37 (0.17) & 0.49 (0.35) \\
CM \citep{zhao2024crossmatch}               & 0.67 (0.18) & 0.00 (0.00) & 0.82 (0.12) & 0.34 (0.23) & 0.00 (0.00) & 0.32 (0.71) \\
CML \citep{wu2024cross}              & \underline{0.70 (0.14)} & 0.44 (0.21) & \textbf{0.84 (0.08)} & \textbf{0.40 (0.23)} & 0.38 (0.17) & \underline{0.51 (0.36)} \\
DyCON  \citep{assefa2025dycon}            & 0.58 (0.15) & 0.40 (0.16) & 0.81 (0.10) & 0.32 (0.23) & 0.31 (0.19) & 0.45 (0.36) \\
MagicNet  \citep{chen2023magicnet}          & 0.64 (0.19) & 0.43 (0.16) & 0.79 (0.06) & 0.34 (0.26) & 0.34 (0.19) & 0.48 (0.34) \\
RegAL (Ours)    & \textbf{0.71 (0.13)} & \textbf{0.47 (0.16)} & \textbf{0.84 (0.10)} & \underline{0.38 (0.21)} & \textbf{0.42 (0.17)} & \textbf{0.53 (0.36)} \\
\bottomrule
\end{tabular}%
}
\end{table}

\begin{table}[H]
\centering
\caption{Comparison of  \textbf{regional} ASD scores ($\downarrow$) of SSL strategies on ProstateX data using 10, 20 and 100 labeled data/
Lower is better. \textbf{Bold} indicates the best performance; \underline{underline} indicates the second-best. \textbf{Note:} N/A indicates an undefined ASD value resulting from the absence of predicted foreground pixels for that anatomical region.}
\label{tab:prostatex_asd_complete}
\resizebox{\textwidth}{!}{%
\begin{tabular}{lcccccc}
\toprule
\textbf{Method} & \textbf{PZ} & \textbf{CZ} & \textbf{TZ} & \textbf{AS} & \textbf{Urethra} & \textbf{Average} \\
\midrule
% --- 10 LABELED ---
\multicolumn{7}{c}{\textbf{\textit{Experiment 1: 10 Labeled Samples}}} \\
\midrule
U-Net (Baseline) & 3.54 (5.11) & \underline{1.71 (1.38)} & 2.98 (12.60) & 1.83 (2.46) & \underline{2.02 (3.73)} & \underline{2.10 (4.09)} \\
BCP \citep{bai2023bidirectional}            & 3.28 (3.39) & 3.04 (4.74) & 2.33 (4.11) & 2.95 (3.79) & 2.90 (2.80) & 2.94 (3.82) \\
CM  \citep{zhao2024crossmatch}            & \underline{1.62 (1.89)} & N/A & \underline{1.84 (1.69)} & 4.25 (4.09) & N/A & 2.19 (3.60) \\
CML \citep{wu2024cross}            & 2.01 (5.21) & 7.02 (19.38) & 2.12 (3.04) & \underline{1.58 (4.20)} & 4.61 (8.16) & 2.74 (7.77) \\
DyCON \citep{assefa2025dycon}           & 3.50 (4.31) & 17.83 (21.94) & 9.28 (4.77) & 2.41 (3.54) & 11.78 (9.78) & 9.01 (10.03) \\
MagicNet \citep{chen2023magicnet}        & 2.96 (2.32) & 2.69 (2.41) & 2.14 (2.61) & 2.70 (2.85) & 3.86 (2.34) & 2.93 (2.80) \\
RegAL (Ours)   & \textbf{1.07 (0.85)} & \textbf{1.44 (0.92)} & \textbf{1.25 (0.85)} & \textbf{1.31 (1.44)} & \textbf{1.76 (1.12)} & \textbf{1.38 (1.18)} \\
\midrule

% --- 20 LABELED ---
\multicolumn{7}{c}{\textbf{\textit{Experiment 2: 20 Labeled Samples}}} \\
\midrule
U-Net (Baseline) & 2.59 (4.62) & 1.50 (1.28) & 13.31 (23.67) & \underline{1.53 (1.58)} & 1.91 (2.60) & 2.05 (4.66) \\
BCP \citep{bai2023bidirectional}            & \underline{1.40 (2.29)} & 2.13 (1.88) & 1.46 (2.07) & 3.03 (4.92) & 1.98 (2.08) & 1.88 (2.61) \\
CM \citep{zhao2024crossmatch}              & 1.61 (2.91) & 2.12 (1.53) & \underline{1.39 (1.45)} & N/A & N/A & \underline{1.77 (1.97)} \\
CML \citep{wu2024cross}              & 1.84 (3.59) & 3.33 (9.31) & 2.14 (3.21) & 3.35 (11.90) & \textbf{1.69 (1.50)} & 2.09 (4.00) \\
DyCON \citep{assefa2025dycon}           & 3.81 (2.75) & \underline{1.37 (2.08)} & 2.71 (1.82) & 1.57 (2.64) & 8.16 (16.12) & 3.14 (4.11) \\
MagicNet \citep{chen2023magicnet}        & 1.88 (1.87) & 1.88 (2.01) & 1.58 (1.40) & 2.71 (2.57) & 2.56 (1.89) & 2.11 (2.01) \\
RegAL (Ours)   & \textbf{0.94 (0.93)} & \textbf{1.31 (0.91)} & \textbf{1.10 (0.69)} & \textbf{1.37 (1.69)} & \underline{1.70 (1.43)} & \textbf{1.29 (1.22)} \\
\midrule

% --- 100 LABELED ---
\multicolumn{7}{c}{\textbf{\textit{Experiment 3: 100 Labeled Samples}}} \\
\midrule
U-Net (Oracle) & 1.11 (1.31) & 1.27 (0.63) & 0.92 (0.77) & \underline{1.31 (1.59)} & \underline{1.35 (1.17)} & 1.23 (1.08) \\
BCP \citep{bai2023bidirectional}             & 1.27 (1.18) & 1.35 (0.93) & 1.06 (0.65) & 2.00 (2.05) & 1.59 (0.85) & 1.42 (1.12) \\
CM \citep{zhao2024crossmatch}              & \underline{1.06 (0.65)} & N/A & 1.07 (1.03) & 1.46 (1.72) & N/A & 1.15 (1.20) \\
CML \citep{wu2024cross}              & \textbf{0.86 (0.39)} & \textbf{1.04 (0.58)} & \underline{0.91 (0.39)} & \textbf{1.07 (0.80)} & \textbf{1.32 (0.82)} & \textbf{1.02 (0.66)} \\
DyCON \citep{assefa2025dycon}           & 1.19 (1.08) & 1.26 (1.07) & 1.26 (0.92) & 1.34 (1.19) & 1.44 (1.04) & 1.30 (1.03) \\
MagicNet \citep{chen2023magicnet}         & 1.51 (1.23) & 1.46 (1.06) & 1.26 (0.77) & 1.69 (1.45) & 1.96 (1.51) & 1.54 (1.34) \\
RegAL (Ours)    & \textbf{0.86 (0.43)} & \underline{1.13 (0.66)} & \textbf{0.86 (0.42)} & 1.40 (1.46) & 1.46 (1.00) & \underline{1.07 (0.84)} \\
\bottomrule
\end{tabular}%
}
\end{table}

\begin{figure*}[t]
    \centering
    \includegraphics[width=\textwidth]{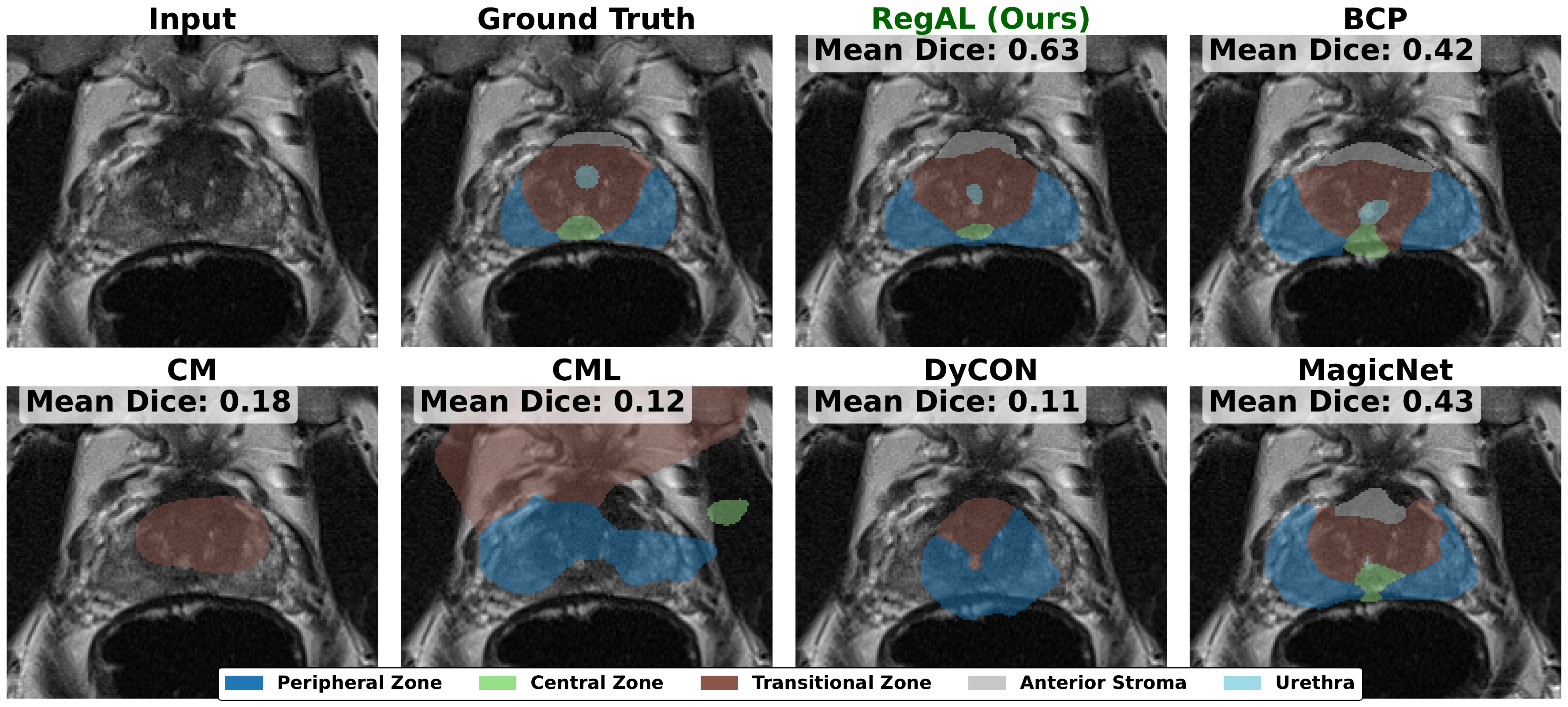}
    \caption{\textbf{Qualitative comparison of SSL strategies on the ProstateX dataset (20 labeled samples).} The first panel shows the cropped Input from a randomly selected T2 MRI focused on the pelvic region. Corresponding 3D Mean Dice scores are displayed within each subplot. RegAL demonstrates superior boundary adherence and anatomical plausibility, better identifying minor structures such as the Urethra and Anterior Stroma where consistency-based baselines fail to generalize.}
    \label{fig:prostate_qualitative}
\end{figure*}

\begin{figure}[!t]
    \centering
    \includegraphics[width=\textwidth]{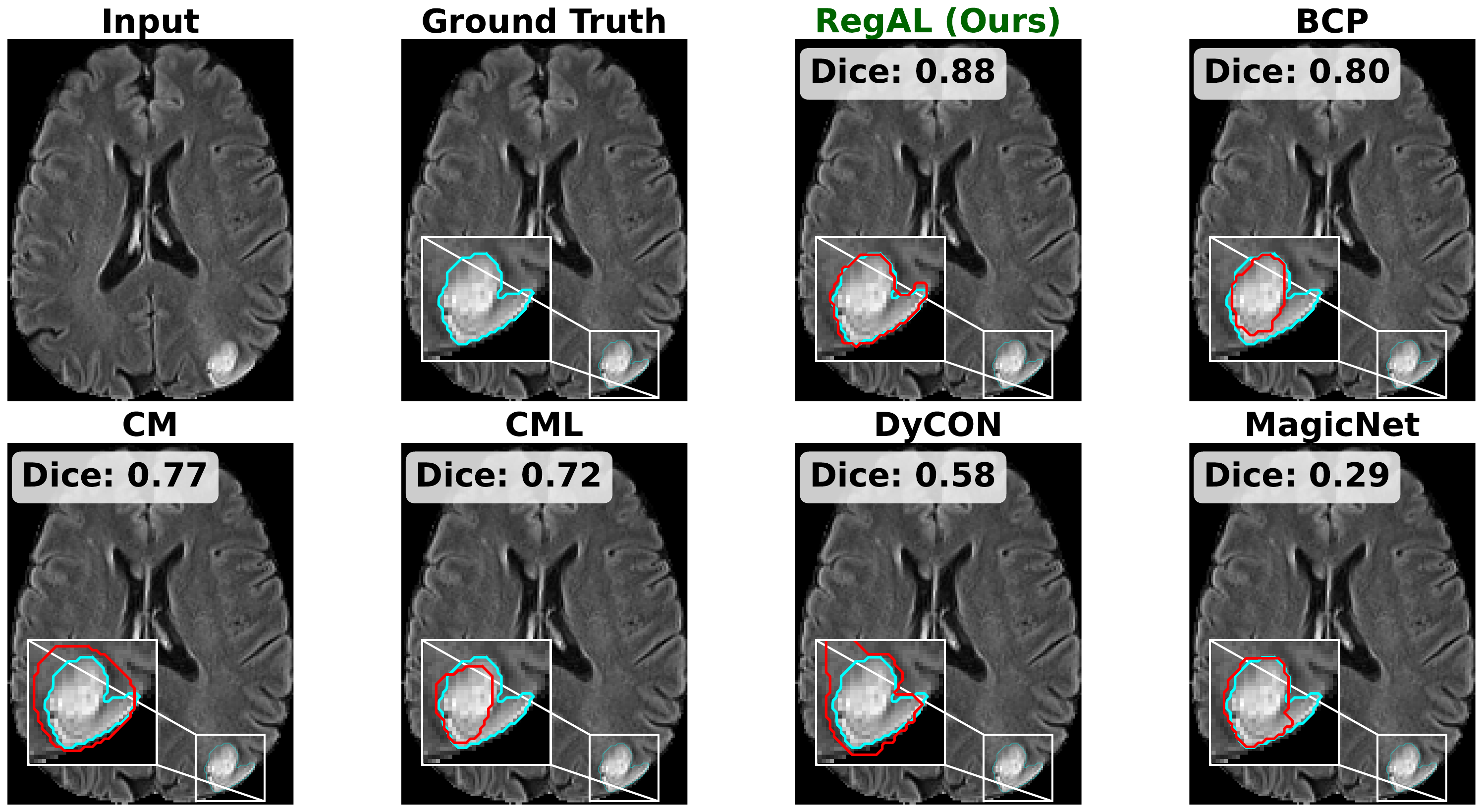}
    \caption{\textbf{Qualitative comparison of SSL strategies on BraTS 2021 (10 labeled samples).} A randomly selected FLAIR is displayed with Ground Truth (cyan) and model predicted segmentations (red). Zoomed insets (white boxes) highlight boundary adherence. 3D Whole Tumor Dice scores are shown in each panel.}
    \label{fig:qualitative_brats}
\end{figure}

Fig.~\ref{fig:prostate_qualitative} and Fig.~\ref{fig:qualitative_brats} provide qualitative segmentation comparisons for ProstateX (20 labeled samples) and BraTS (10 labeled samples), respectively. Complete Urethra/AS dropout in Fig.~\ref{fig:prostate_qualitative} is precisely the failure mode addressed by RegAL's topology-aware Pareto objective and registration-guided augmentation.

\clearpage

On dHCP, Tables~\ref{tab:dhcp_dice_only} and \ref{tab:dhcp_asd_only} disaggregate the summary Dice and ASD reported in the main paper (Table~6) into individual tissue classes, enabling assessment of per-structure gains.
RegAL's largest advantages occur on topologically vulnerable structures --- Ventricles and Hippocampus --- where consistency-based baselines frequently collapse to near-zero predictions in the extreme 5-label regime.

\begin{table}[H]
\centering
\caption{Comparison of Dice scores ($\uparrow$) of SSL strategies using 5 and 10 labeled samples on dHCP data. \textbf{Abbreviations:} Vent (Ventricle), Cer (Cerebellum), BS (Brainstem), Hippo (Hippocampus). Higher is better. \textbf{Bold} indicates best; \underline{underline} indicates second-best. Values are reported as Median (IQR).}
\label{tab:dhcp_dice_only}
\setlength{\tabcolsep}{3pt}
\resizebox{\textwidth}{!}{%
\begin{tabular}{lccccccccc}
\toprule
\textbf{Method} & \textbf{CSF} & \textbf{CGM} & \textbf{WM} & \textbf{Vent} & \textbf{Cer} & \textbf{DGM} & \textbf{BS} & \textbf{Hippo} & \textbf{Avg} \\
\midrule
% --- DICE SECTION 1: 5 SAMPLES ---
\multicolumn{10}{c}{\textbf{\textit{Experiment 1: 5 Labeled Samples}}} \\
\midrule
U-Net (Baseline) & 0.83 (0.04) & 0.85 (0.03) & 0.89 (0.03) & 0.58 (0.15) & 0.85 (0.12) & 0.86 (0.04) & 0.88 (0.08) & 0.53 (0.14) & 0.84 (0.19) \\
\midrule
BCP \citep{bai2023bidirectional}             & 0.83 (0.03) & \textbf{0.87 (0.02)} & \textbf{0.91 (0.03)} & 0.75 (0.08) & \underline{0.94 (0.02)} & \underline{0.90 (0.02)} & 0.90 (0.04) & 0.69 (0.08) & 0.87 (0.11) \\
CM  \citep{zhao2024crossmatch}             & \underline{0.84 (0.03)} & 0.86 (0.02) & 0.89 (0.02) & 0.70 (0.10) & \textbf{0.95 (0.01)} & \underline{0.90 (0.02)} & \underline{0.91 (0.02)} & 0.39 (0.03) & 0.87 (0.12) \\
CML  \citep{wu2024cross}           & \underline{0.84 (0.03)} & \textbf{0.87 (0.02)} & \textbf{0.91 (0.03)} & \underline{0.77 (0.08)} & \textbf{0.95 (0.01)} & \underline{0.91 (0.02)} & 0.88 (0.04) & 0.69 (0.07) & 0.87 (0.11) \\
DyCON   \citep{assefa2025dycon}        & 0.73 (0.03) & 0.74 (0.04) & 0.81 (0.06) & 0.27 (0.10) & 0.82 (0.08) & 0.76 (0.06) & 0.68 (0.15) & 0.30 (0.21) & 0.73 (0.25) \\
MagicNet \citep{chen2023magicnet}       & 0.71 (0.05) & 0.79 (0.03) & 0.80 (0.05) & 0.33 (0.12) & 0.69 (0.14) & 0.21 (0.16) & 0.16 (0.09) & 0.10 (0.12) & 0.45 (0.56) \\
\textbf{RegAL (Ours)} & 0.83 (0.03) & \textbf{0.87 (0.03)} & \textbf{0.92 (0.03)} & \textbf{0.83 (0.06)} & \textbf{0.95 (0.04)} & \textbf{0.93 (0.02)} & \textbf{0.93 (0.02)} & \textbf{0.81 (0.06)} & \textbf{0.88 (0.10)} \\
\midrule

% --- DICE SECTION 2: 10 SAMPLES ---
\multicolumn{10}{c}{\textbf{\textit{Experiment 2: 10 Labeled Samples}}} \\
\midrule
U-Net (Baseline) & 0.84 (0.03) & 0.86 (0.03) & 0.90 (0.03) & 0.65 (0.13) & 0.90 (0.07) & 0.88 (0.03) & 0.85 (0.08) & 0.64 (0.14) & 0.85 (0.14) \\
\midrule
BCP   \citep{bai2023bidirectional}           & 0.84 (0.03) & 0.87 (0.02) & 0.91 (0.03) & 0.78 (0.07) & 0.94 (0.02) & 0.90 (0.02) & 0.90 (0.04) & 0.72 (0.08) & 0.88 (0.10) \\
CM   \citep{zhao2024crossmatch}           & \textbf{0.85 (0.03)} & 0.87 (0.02) & 0.91 (0.02) & \textbf{0.84 (0.05)} & \textbf{0.95 (0.01)} & \underline{0.93 (0.01)} & \textbf{0.93 (0.02)} & \textbf{0.82 (0.04)} & \textbf{0.89 (0.08)} \\
CML  \citep{wu2024cross}           & \textbf{0.85 (0.03)} & \textbf{0.88 (0.02)} & \textbf{0.92 (0.03)} & 0.80 (0.06) & \textbf{0.95 (0.01)} & \underline{0.93 (0.01)} & 0.90 (0.04) & 0.75 (0.06) & 0.88 (0.10) \\
DyCON  \citep{assefa2025dycon}         & 0.77 (0.03) & 0.79 (0.04) & 0.85 (0.05) & 0.58 (0.09) & 0.85 (0.07) & 0.81 (0.04) & 0.76 (0.11) & 0.25 (0.17) & 0.78 (0.18) \\
MagicNet  \citep{chen2023magicnet}      & 0.78 (0.04) & 0.83 (0.03) & 0.84 (0.04) & 0.27 (0.12) & 0.85 (0.09) & 0.47 (0.19) & 0.50 (0.22) & 0.20 (0.27) & 0.67 (0.46) \\
\textbf{RegAL (Ours)} & \textbf{0.85 (0.03)} & \textbf{0.88 (0.03)} & \textbf{0.92 (0.03)} & \underline{0.83 (0.05)} & \textbf{0.95 (0.02)} & \textbf{0.94 (0.01)} & \textbf{0.93 (0.02)} & \underline{0.80 (0.06)} & \textbf{0.89 (0.10)} \\
\bottomrule
\end{tabular}%
}
\end{table}

Table~\ref{tab:dhcp_asd_only} reports the corresponding boundary adherence (ASD, in mm) for the same experimental conditions.
RegAL achieves the lowest ASD on six of eight tissue classes at both label budgets, confirming that Dice improvements are accompanied by tighter boundary delineation rather than inflated by voxel overlap alone.

\begin{table}[H]
\centering
\caption{Comparison of ASD scores ($\downarrow$) of SSL strategies using 5 and 10 labeled samples on dHCP data. \textbf{Abbreviations:} Vent (Ventricle), Cer (Cerebellum), BS (Brainstem), Hippo (Hippocampus). Lower is better. \textbf{Bold} indicates best; \underline{underline} indicates second-best. Values are reported as Median (IQR).}
\label{tab:dhcp_asd_only}
\setlength{\tabcolsep}{3pt}
\resizebox{\textwidth}{!}{%
\begin{tabular}{lccccccccc}
\toprule
\textbf{Method} & \textbf{CSF} & \textbf{CGM} & \textbf{WM} & \textbf{Vent} & \textbf{Cer} & \textbf{DGM} & \textbf{BS} & \textbf{Hippo} & \textbf{Avg} \\
\midrule
% --- ASD SECTION 1: 5 SAMPLES ---
\multicolumn{10}{c}{\textbf{\textit{Experiment 1: 5 Labeled Samples}}} \\
\midrule
U-Net (Baseline) & 0.40 (0.12) & 0.53 (0.18) & 0.58 (0.26) & 0.87 (0.59) & 3.96 (2.45) & 1.26 (0.55) & 1.39 (1.44) & 2.83 (2.68) & 0.97 (1.55) \\
\midrule
BCP  \citep{bai2023bidirectional}             & 0.31 (0.08) & 0.37 (0.08) & 0.40 (0.08) & 0.63 (0.23) & 0.74 (0.44) & 1.00 (0.40) & 0.77 (0.58) & 1.10 (0.84) & 0.60 (0.55) \\
CM   \citep{zhao2024crossmatch}             & \textbf{0.28 (0.04)} & 0.34 (0.07) & 0.42 (0.09) & 1.54 (0.81) & 0.77 (0.59) & 1.52 (0.80) & 0.54 (0.31) & 6.15 (1.30) & 0.64 (1.24) \\
CML  \citep{wu2024cross}           & 0.31 (0.09) & 0.36 (0.07) & 0.37 (0.08) & \textbf{0.46 (0.15)} & \textbf{0.42 (0.19)} & \textbf{0.62 (0.17)} & 0.55 (0.30) & 0.67 (0.18) & 0.46 (0.25) \\
DyCON  \citep{assefa2025dycon}         & 0.65 (0.14) & 0.61 (0.15) & 0.83 (0.16) & 1.43 (0.70) & 2.71 (1.24) & 3.14 (1.09) & 6.48 (2.32) & 2.35 (1.31) & 1.66 (2.39) \\
MagicNet \citep{chen2023magicnet}       & 0.54 (0.12) & 0.53 (0.09) & 0.90 (0.12) & 5.79 (1.22) & 8.09 (3.02) & 7.83 (1.99) & 9.83 (3.58) & 1.94 (1.58) & 4.41 (6.96) \\
\textbf{RegAL (Ours)} & \textbf{0.28 (0.07)} & \textbf{0.32 (0.07)} & \textbf{0.35 (0.12)} & \underline{0.48 (0.27)} & \underline{0.62 (0.55)} & \underline{0.74 (0.80)} & \textbf{0.38 (0.19)} & \textbf{0.51 (0.17)} & \textbf{0.43 (0.28)} \\
\midrule

% --- ASD SECTION 2: 10 SAMPLES ---
\multicolumn{10}{c}{\textbf{\textit{Experiment 2: 10 Labeled Samples}}} \\
\midrule
U-Net (Baseline) & 0.38 (0.13) & 0.39 (0.10) & 0.44 (0.12) & 0.83 (0.51) & 1.92 (1.82) & 0.98 (0.41) & 1.80 (1.80) & 2.70 (2.74) & 0.81 (1.22) \\
\midrule
BCP  \citep{bai2023bidirectional}             & 0.30 (0.07) & 0.34 (0.08) & 0.37 (0.09) & 0.55 (0.17) & 0.85 (0.57) & 0.98 (0.43) & 0.60 (0.62) & 1.36 (1.16) & 0.56 (0.57) \\
CM   \citep{zhao2024crossmatch}             & \underline{0.27 (0.04)} & 0.33 (0.07) & 0.36 (0.08) & 0.50 (0.26) & 0.53 (0.32) & 0.72 (0.26) & 0.41 (0.23) & 0.77 (0.46) & 0.44 (0.34) \\
CML  \citep{wu2024cross}           & 0.29 (0.08) & 0.35 (0.08) & 0.36 (0.09) & \textbf{0.39 (0.12)} & \underline{0.48 (0.26)} & 0.54 (0.12) & 0.58 (0.43) & 0.57 (0.16) & \underline{0.43 (0.21)} \\
DyCON  \citep{assefa2025dycon}         & 0.55 (0.12) & 0.54 (0.14) & 0.63 (0.13) & 2.80 (1.00) & 1.98 (0.72) & 2.09 (0.78) & 3.18 (1.65) & 1.44 (1.01) & 1.57 (1.86) \\
MagicNet  \citep{chen2023magicnet}      & 0.48 (0.12) & 0.49 (0.09) & 0.73 (0.11) & 1.80 (1.10) & 2.21 (1.68) & 5.94 (1.77) & 5.88 (4.04) & 11.37 (7.58) & 1.91 (5.17) \\
\textbf{RegAL (Ours)} & \textbf{0.26 (0.05)} & \textbf{0.32 (0.07)} & \textbf{0.32 (0.11)} & \underline{0.40 (0.17)} & \textbf{0.47 (0.38)} & \textbf{0.48 (0.14)} & \textbf{0.37 (0.20)} & \textbf{0.51 (0.17)} & \textbf{0.39 (0.21)} \\
\bottomrule
\end{tabular}%
}
\end{table}

\begin{figure*}[t]
    \centering
    \includegraphics[width=\textwidth]{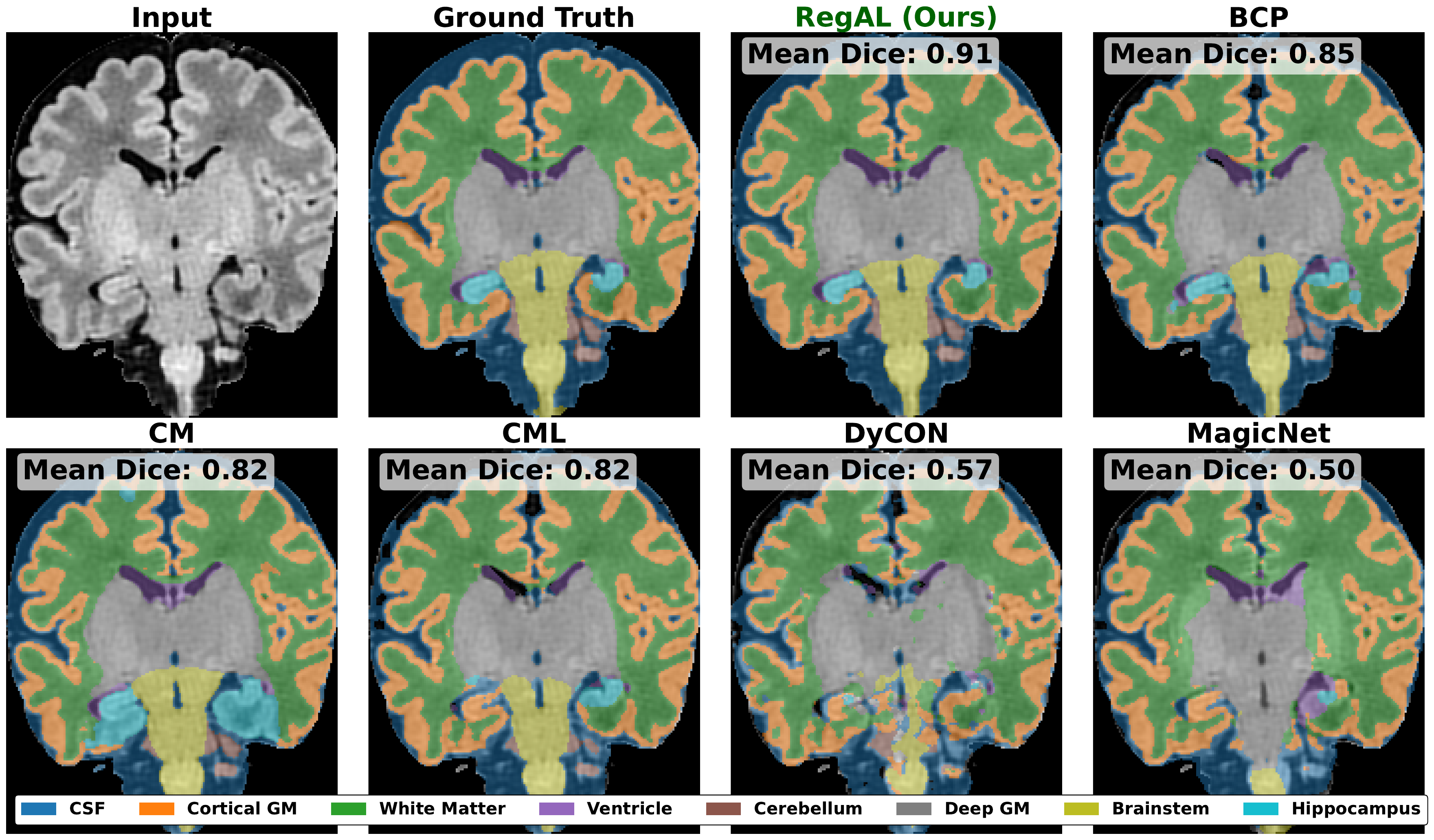}
    \caption{\textbf{Qualitative comparison of SSL strategies on the dHCP dataset using an ultra-low budget of 5 labeled samples.} A randomly selected T1-weighted MRI is displayed alongside the Ground Truth and predicted segmentations from state-of-the-art SSL baselines. Anatomical structures are color-coded (see legend) into eight tissue classes. Mean 3D Dice scores are provided for each method. While baselines such as MagicNet and DyCON exhibit significant label noise and structural fragmentation, RegAL (Ours) maintains high topological integrity, particularly in the hippocampus.}
    \label{fig:dhcp_qualitative}
\end{figure*}

Fig.~\ref{fig:dhcp_qualitative} shows a representative qualitative comparison on dHCP (5 labeled samples). Fragmentation of the Hippocampus and Ventricles is precisely the failure mode addressed by RegAL's topology-aware Pareto objective and registration-guided augmentation.

\clearpage

% === end hidden augmentation ablation ===

\bibliographystyle{cas-model2-names}
\bibliography{cas-refs}

\end{document}